\documentclass[acmsmall]{acmart} 

\usepackage{algorithm}
\usepackage{algorithmic}
\usepackage{amsmath}
\usepackage{graphicx}
\usepackage{multirow}
\usepackage{booktabs}
\usepackage{subcaption}
\usepackage{float}
\usepackage{bm}
\usepackage{listings}
\usepackage{xcolor}
\usepackage{caption}

\definecolor{codegray}{rgb}{0.99, 0.99, 0.99}
\definecolor{codegreen}{HTML}{089900}

\lstdefinestyle{mystyle}{
    backgroundcolor=\color{codegray},   
    commentstyle=\color{codegreen},
    keywordstyle=\color{blue},
    numberstyle=\tiny\color{gray},
    stringstyle=\color{purple},
    basicstyle=\ttfamily\footnotesize,
    breakatwhitespace=false,         
    breaklines=true,                 
    captionpos=b,                    
    keepspaces=true,                 
    numbers=none,                    
    numbersep=8pt,                  
    showspaces=false,                
    showstringspaces=false,
    showtabs=false,                  
    tabsize=2,
    frame=single,                   
    framesep=1pt,
    rulecolor=\color{black},
    framerule=1pt,
}

\newif\iffinal

\finaltrue

\iffinal
  
  \newcommand{\add}[1]{#1}
\else
  
  \newcommand{\add}[1]{{\textcolor{blue}{#1}}}
\fi

\AtBeginDocument{}

\setcopyright{acmlicensed}
\copyrightyear{2024}
\acmYear{2024}
\acmDOI{XXXXXXX.XXXXXXX}

\begin{document}

\title{TensorNEAT: A GPU-accelerated Library for NeuroEvolution of Augmenting Topologies}

\author{Lishuang Wang}
\orcid{0009-0009-2374-8699}
\email{wanglishuang22@gmail.com}
\affiliation{%
  \institution{Southern University of Science and Technology}
  \city{Shenzhen}
  \state{Guangdong}
  \country{China}
  \postcode{518055}
}

\author{Mengfei Zhao}
\orcid{0009-0002-1653-8553}
\email{12211819@mail.sustech.edu.cn}
\affiliation{%
  \institution{Southern University of Science and Technology}
  \city{Shenzhen}
  \state{Guangdong}
  \country{China}
  \postcode{518055}
}

\author{Enyu Liu}
\orcid{0009-0008-5645-8578}
\email{12210417@mail.sustech.edu.cn}
\affiliation{%
  \institution{Southern University of Science and Technology}
  \city{Shenzhen}
  \state{Guangdong}
  \country{China}
  \postcode{518055}
}

\author{Kebin Sun}
\orcid{0009-0008-9213-7835}
\email{sunkebin.cn@gmail.com}
\affiliation{%
  \institution{Southern University of Science and Technology}
  \city{Shenzhen}
  \state{Guangdong}
  \country{China}
  \postcode{518055}
}

\author{Ran Cheng}
\orcid{0000-0001-9410-8263}
\authornote{Corresponding Author}
\email{ranchengcn@gmail.com}
\affiliation{%
    \institution{The Hong Kong Polytechnic University}
    \state{Hong Kong SAR}
    \country{China}
}


\begin{abstract}

The NeuroEvolution of Augmenting Topologies (NEAT) algorithm has received considerable recognition in the field of neuroevolution. 
Its effectiveness is derived from initiating with simple networks and incrementally evolving both their topologies and weights.
Although its capability across various challenges is evident, the algorithm's computational efficiency remains an impediment, limiting its scalability potential.  
To address these limitations, this paper introduces TensorNEAT, a GPU-accelerated library that applies tensorization to the NEAT algorithm. 
Tensorization reformulates NEAT's diverse network topologies and operations into uniformly shaped tensors, enabling efficient parallel execution across entire populations. 
TensorNEAT is built upon JAX, leveraging automatic function vectorization and hardware acceleration to significantly enhance computational efficiency. 
In addition to NEAT, the library supports variants such as CPPN and HyperNEAT, and integrates with benchmark environments like Gym, Brax, and gymnax. 
Experimental evaluations across various robotic control environments in Brax demonstrate that TensorNEAT delivers up to 500x speedups compared to existing implementations, such as NEAT-Python. 
The source code for TensorNEAT is publicly available at: \url{https://github.com/EMI-Group/tensorneat}.
\end{abstract}




\maketitle

\section{Introduction}
Neuroevolution has emerged as a distinct branch within the field of artificial intelligence (AI). Unlike the common approach in machine learning that uses stochastic gradient descent, neuroevolution employs evolutionary algorithms for network optimization. This method not only optimizes parameters but also improves more complex aspects such as activation functions, hyperparameters, and the overall network architecture. Furthermore, neuroevolution offers a distinct advantage over traditional machine learning methods, which often converge to a single solution by relying on gradient-based techniques; instead, neuroevolution maintains a diverse population of solutions throughout its search process, allowing it to explore a broader range of possibilities \cite{stanley2019designing}. This characteristic of sustaining a variety of potential solutions makes neuroevolution particularly well-suited for problems with non-stationary environments or open-ended objectives, where the optimal solution may change over time or where there may be multiple viable solutions \cite{lehman2011evolving, mouret2015illuminating}. These qualities not only emphasize neuroevolution's capacity for exploration but also highlight its potential for creative problem solving and its ability to tackle challenges that require adaptability and long-term innovation.

The NeuroEvolution of Augmenting Topologies (NEAT) \cite{stanley2002evolving} is well-recognized in the neuroevolution literature, particularly for its pioneering approach to evolving neural networks with both topological and weight diversity. Since its introduction in 2002, NEAT has demonstrated versatility across a wide range of domains, including game AI \cite{stanley2006real, pham2018playing}, where it has been employed for evolving adaptive agents, robotics \cite{silva2012odneat, auerbach2011evolving}, enabling the design of efficient control strategies, and self-driving systems \cite{yuksel2018agent}, contributing to advancements in autonomous navigation. The original NEAT algorithm provided a solid foundation, and subsequent research has continued to expand its capabilities, leveraging its core principles to explore new areas. For example, extensions like HyperNEAT \cite{stanley2009hypercube} and ES-HyperNEAT \cite{risi2010evolving} adopted indirect encodings to evolve large-scale networks with regular patterns, while DeepNEAT and CoDeepNEAT \cite{miikkulainen2019evolving} integrated gradient-based methods to explore deeper neural architectures for more complex tasks. Recently, the RankNEAT algorithm \cite{rankneat} has applied NEAT principles to preference learning, where it optimizes networks based on subjective data, expanding NEAT's application into more nuanced, human-centered decision-making problems. The continued development of these variations not only underscores NEAT's enduring relevance but also showcases its adaptability in addressing diverse and evolving challenges within neuroevolution, making it a cornerstone in the field's ongoing innovation.

Over the past decade, the role of GPU acceleration has been a key factor in driving the rapid progress of artificial intelligence, particularly within deep learning. As models become increasingly complex, requiring the processing of hundreds of billions of parameters, such as in the case of modern large language models \cite{brown2020language}, hardware acceleration has proven crucial for handling the associated computational demands. GPUs, when employed for tasks like inference and back-propagation, have not only contributed to substantial reductions in training time but also enabled more scalable model architectures that can learn from vast datasets. Similarly, in the neuroevolution domain, there has been a growing effort to leverage the power of GPUs to achieve similar performance improvements. The use of frameworks like JAX~\cite{frostig2018compiling} has spurred developments in GPU-accelerated neuroevolution algorithms, with notable contributions from projects such as EvoJAX \cite{tang2022evojax}, evosax \cite{lange2022evosax}, and EvoX \cite{huang2023evox}. These works aim to exploit the parallel processing capabilities of GPUs to significantly reduce the computational time required for evolutionary algorithms, which is particularly advantageous when dealing with large-scale problems or managing sizable populations. As a result, GPU acceleration is becoming increasingly integrated into neuroevolution workflows, paving the way for more efficient and scalable solutions in both research and practical applications.

Despite the rapid emergence of GPU-accelerated libraries across various machine learning and evolutionary computation frameworks, NEAT has remained largely underdeveloped in this regard. Our analysis revealed that while the NEAT algorithm has been implemented in several programming languages~\cite{McIntyre_neat-python, MultiNEAT, MonopolyNEAT}, very few of these implementations leverage the computational power of GPUs to enhance performance. On the rare occasions where GPUs are utilized~\cite{pytorchneat}, the focus has primarily been on accelerating only the network inference aspect of NEAT, neglecting other vital components such as the evolutionary search and network mutation processes, which are critical to the algorithm's success. Moreover, even in these GPU-accelerated implementations, the potential of parallel computation is not fully realized, as essential operations like fitness evaluation and mutation are still executed sequentially, preventing the framework from harnessing the full power of modern GPUs. This underutilization can be attributed to the inherent complexity of NEAT, which continuously evolves network topologies throughout its execution. This dynamic and unpredictable nature of topology evolution poses significant challenges for an efficient and scalable GPU implementation, as it complicates the parallelization of key operations that are essential for NEAT's performance.

To address the performance limitations of traditional NEAT implementations, we introduce TensorNEAT, a GPU-accelerated library of the NEAT algorithm. TensorNEAT leverages a novel tensorization strategy that converts networks of diverse topologies into uniformly shaped tensors, enabling parallel execution of operations across the entire population. This approach ensures that the algorithm's performance scales efficiently with the complexity of the task and the size of the population. By utilizing the JAX framework, TensorNEAT automatically benefits from GPU acceleration without requiring manual configuration or specialized hardware knowledge. When compared to widely-used open-source NEAT implementations, TensorNEAT demonstrates speedups of up to 500x, significantly reducing the time required for evolving neural networks. 
Overall, our contributions are summarized as follows.

\begin{itemize} 
    \item We propose a tensorization method, which enables the transformation of networks with various topologies and their associated operations in the NEAT algorithm into uniformly structured tensors for tensor computation. This method allows operations within the NEAT algorithm to be executed in parallel across the entire population, thereby enhancing the efficiency of the process.

    \item We develop TensorNEAT, a GPU-accelerated NEAT library based on JAX, characterized by high efficiency, flexible adaptability, and rich capabilities. TensorNEAT supports full GPU acceleration of representative NEAT algorithms, including the original NEAT algorithm, CPPN~\cite{stanley2007compositional}, and HyperNEAT~\cite{stanley2009hypercube}. It also provides seamless interfaces with advanced control benchmarks, including Brax~\cite{freeman2021brax} and Gymnax~\cite{lange2022gymnax}, featuring GPU-accelerated environments with various classical control and robotics control tasks.

    \item We assessed TensorNEAT's performance in a spectrum of complex robotics control tasks, benchmarked against the NEAT-Python library~\cite{McIntyre_neat-python}. The results show that TensorNEAT significantly outperforms in terms of execution speed, especially under high computational demands and across various population sizes and network scales.

\end{itemize}

This paper is an extension of its prior conference version~\cite{tensorneat_gecco}. In particular, (a) we expanded the Implementation of TensorNEAT section, providing a more detailed description of the TensorNEAT library, including the introduction of its visualization capabilities; (b) we introduced multi-GPU support for TensorNEAT, and conducted experiments to evaluate the performance benefits of this feature; (c) we restructured the paper's content and organization, including a title change, the enrichment of various sections, and improvements to the figures and visuals throughout the manuscript; \add{and (d) we supplemented the experimental evaluation by testing TensorNEAT's performance under different population size settings and comparing it with other GPU-accelerated EC algorithm libraries.} 

\section{Background}

\subsection{NeuroEvolution of Augmenting Topologies}

\begin{algorithm}[t]
    \caption{Main Process of the NEAT algorithm}
    \label{alg:neat}
    \begin{algorithmic}
    \REQUIRE $P$ (population size), $I$ (number of input nodes), $O$ (number of output nodes), $f_{\text{target}}$ (target fitness value), $G$ (maximum number of generations) 
    \ENSURE $best$
    \STATE $pop \leftarrow$ initialize $P$ networks with $I$, $O$
    \FOR{$g = 1$ to $G$}
        \STATE $fit \leftarrow$ evaluate fitness values of $Pop$
        \IF{$\max(fit) \geq f_{\text{target}}$}
            \STATE \textbf{break}
        \ENDIF
        \STATE $species \leftarrow$ divide $pop$ by distances between networks
        
        \STATE $pop^* \leftarrow \{\}$ 
        \FOR{$s$ in $species$}
            \STATE $c \leftarrow$ determine the number of new individuals by $fit$
            \STATE $s \leftarrow$ generate $c$ networks using crossover and mutation
            \STATE $pop^* \leftarrow pop^* \cup s$
        \ENDFOR
        \STATE $pop \leftarrow pop^*$
    \ENDFOR
    \STATE \textbf{return} $pop[\arg\max(fit)]$
    \end{algorithmic}
\end{algorithm}

Introduced by Kenneth O. Stanley and Risto Miikkulainen in 2002, the NeuroEvolution of Augmenting Topologies (NEAT) algorithm~\cite{stanley2002evolving} represents a novel approach in neuroevolution. 
The NEAT algorithm manages a range of neural networks and simultaneously optimizes their topologies and weights to identify the most effective networks tailored for designated tasks. Algorithm~\ref{alg:neat} outlines NEAT's core procedure. Starting with a population of simple neural networks, it iterates through evolutionary cycles of species formation, fitness evaluation, and genetic operations. This process dynamically refines the networks until it achieves desired fitness levels or reaches a generational cap, ultimately yielding the optimal network structure.

Setting itself apart from alternative neuroevolution algorithms, NEAT employs three distinctive techniques:
\begin{itemize}
    \item \textbf{Incremental Topological Expansion}: The NEAT algorithm begins its evolutionary process by initializing networks with a minimal configuration, typically consisting of just a single hidden node connecting inputs to outputs. This minimalist starting point is crucial for reducing complexity at the outset, allowing the algorithm to explore the problem space with simpler models before gradually introducing complexity. As evolution proceeds, the algorithm systematically adds new nodes and connections, incrementally expanding the network's topology. This controlled growth not only helps in maintaining a manageable search space but also allows the network to evolve in a more directed manner, progressively building toward more intricate and capable structures. By expanding the topology in stages, NEAT enables a balance between exploration and exploitation, ensuring that simpler network architectures are initially favored while providing the flexibility to evolve into more sophisticated configurations as required. This incremental approach proves especially effective for solving complex problems, as it starts with streamlined architectures and organically grows into larger, more specialized networks as the evolutionary process advances.
    \item \textbf{Historical Markers for Nodes}: Each node in a NEAT network is tagged with a unique historical marker. During the crossover process in the NEAT algorithm, only nodes with identical markers are combined. This method adeptly navigates the challenges of combining networks that possess different topological configurations. By using historical markers, NEAT preserves important structural innovations, ensuring that new nodes and connections are not arbitrarily lost during evolution. This approach promotes stable and consistent network evolution, even as topologies grow more complex.
    \item \textbf{Species-based Population Segmentation}: NEAT categorizes its entire population into species based on genetic similarity, ensuring that networks with common traits evolve together. Conventional genetic procedures, such as selection, mutation, and crossover, are executed independently within each species. This approach serves multiple purposes: it protects promising, newly-formed network structures from being prematurely eliminated by more established competitors, allowing them time to develop and improve. Additionally, species-based segmentation fosters diversity by enabling different topological variations to evolve in parallel, increasing the variety of solutions explored during evolution. This helps prevent premature convergence and promotes a broader exploration of the search space, improving the chances of discovering optimal network configurations.
\end{itemize}

\begin{figure}[b]
    \centering
    \includegraphics[width=0.6\columnwidth]{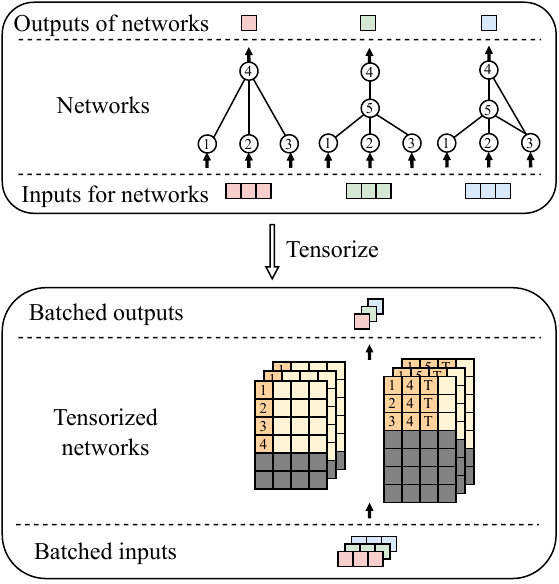}
    \caption{Illustration of the fundamental acceleration principle underlying our tensorization method. In traditional network computations, each network individually processes its inputs. 
    By contrast, with our tensorization method, a single computation suffices to derive the batched output for all networks.
    }
    \label{fig:pop_forward}
\end{figure}

\subsection{NEAT Variants}

Since its inception, the NEAT algorithm has undergone significant evolution, leading to the emergence of several innovative variants, each addressing unique challenges and applications. One of the most notable advancements was the introduction of HyperNEAT~\cite{stanley2009hypercube}, which employed an indirect encoding scheme to enable the efficient generation and handling of large-scale neural networks. By leveraging the inherent geometric regularities found in the problem space, HyperNEAT was able to create intricate network topologies suited for tasks requiring high representational capacity. Building upon this foundation, ES-HyperNEAT~\cite{risi2010evolving} extended the principles of HyperNEAT by incorporating more sophisticated techniques for evolving network structures. ES-HyperNEAT not only retained the scalability benefits of its predecessor but also introduced mechanisms that improved adaptability to complex, high-dimensional problem spaces, thereby broadening the range of tasks for which neuroevolution could be effectively applied. This progression reflects the continuous effort within the neuroevolution community to push the boundaries of what can be achieved with evolving neural architectures, as each variant strives to address the increasing demands of modern applications.

The development of DeepNEAT and CoDeepNEAT~\cite{miikkulainen2019evolving} marked a pivotal advance in integrating neuroevolution with deep learning methodologies, significantly broadening the scope and capabilities of these systems. DeepNEAT took the foundational principles of the NEAT algorithm and extended them into the domain of deep neural networks, allowing for the evolution of more complex, layered architectures that could be fine-tuned using gradient descent. This integration facilitated the exploration of both structure and parameter optimization, giving rise to more sophisticated models that could adapt to a wide range of tasks. CoDeepNEAT further evolved this idea by introducing a co-evolutionary framework, which enabled the simultaneous optimization of both network topologies and their components, such as activation functions or layer types. This dual optimization approach not only increased the flexibility of the models but also enhanced their performance in tasks requiring hierarchical learning and multi-level feature extraction. The combination of evolutionary strategies with gradient-based methods proved to be a powerful tool for discovering novel neural architectures, pushing the boundaries of what was achievable in terms of both accuracy and efficiency in deep learning, and opening new possibilities for more complex, adaptive systems in areas like automated feature learning and the construction of modular, scalable networks.

RankNEAT~\cite{rankneat} is another variant of the NEAT algorithm that leverages neuroevolution to overcome the limitations typically associated with Stochastic Gradient Descent (SGD), particularly in terms of preventing overfitting during network optimization. Unlike SGD, which can struggle with local minima and the complexities of non-convex loss landscapes, RankNEAT explores a broader search space by evolving architectures rather than relying solely on gradient-based updates. \add{This approach has proven highly effective in affective computing, outperforming traditional ranking algorithms such as RankNet, particularly in predicting annotated player arousal from game footage across three diverse games.} RankNEAT's success is particularly evident in tasks involving the analysis of subjective data, such as assessing player arousal from game footage. Its ability to better capture and interpret the nuances of emotional responses makes it a powerful tool in this domain, demonstrating its potential for broader applications where subjective, human-centered data needs to be processed in an adaptive and efficient manner. Furthermore, its robustness in handling non-linear and noisy data suggests that RankNEAT could be adapted for other complex domains, further expanding its utility beyond just affective computing.

Together, these variants of NEAT demonstrate the algorithm's versatility and its ability to continuously adapt to the growing demands of neural network design, application, and optimization. By supporting diverse architectures and a range of problem domains, NEAT has proven effective in evolving both small-scale and large-scale neural networks with varying degrees of complexity.

\begin{figure*}[tbp]
    \centering
    \includegraphics[width=0.98\textwidth]{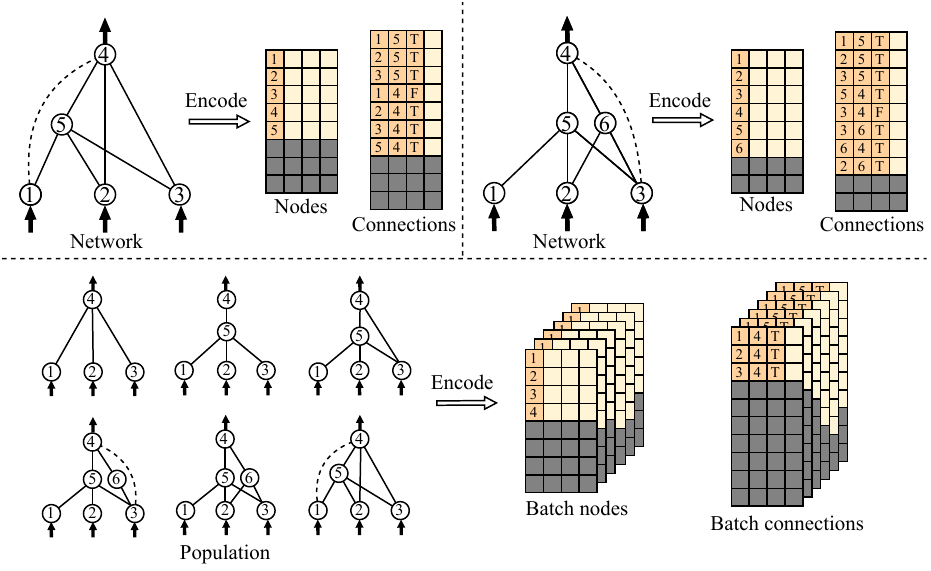}
    \caption{Illustration of the network encoding process. 
    In our method, networks with varying topological structures are transformed into uniformly shaped tensors, enabling the representation of the entire network population as batched tensors. 
    The orange grids symbolize attributes specific to the NEAT algorithm, such as historical markers and enabled flags. 
    The yellow grids denote the attributes of the network's nodes and connections, including biases, weights, and activation functions. 
    The gray grids represent sections filled with \texttt{NaN} to ensure consistent tensor shapes.}
    \label{fig:pop_encoding}
\end{figure*} 

\subsection{NEAT Libraries}
Over the past two decades, the research community has seen the development of various NEAT libraries, including NEAT-Python~\cite{McIntyre_neat-python}, MultiNEAT~\cite{MultiNEAT}, as well as \\ MonopolyNEAT~\cite{MonopolyNEAT}, each contributing to the growing interest in evolving neural networks. Among these, NEAT-Python \cite{McIntyre_neat-python} has emerged as the most widely recognized open-source implementation of NEAT, boasting over 1,200 GitHub stars and serving as the foundation for numerous academic studies and practical applications of the NEAT algorithm \cite{rankneat, gao2021neat, sarti2021neat}. The popularity of NEAT-Python is largely due to its accessibility, ease of use, and robust documentation, which have allowed researchers and developers to build upon its codebase for a wide range of evolutionary computing experiments.

Like many NEAT implementations, NEAT-Python leverages object-oriented programming (OOP) paradigm, where key components such as populations, species, genomes, and genes are encapsulated as objects. This OOP approach fosters code modularity and clarity, making it easier for newcomers to understand the underlying mechanisms of NEAT, while also simplifying debugging and extension of the code. However, this method of representing NEAT components as objects introduces additional memory and computational overhead, particularly when scaling to large population sizes or more complex problem domains. The need to maintain and manipulate large numbers of objects can significantly increase processing time, limiting the efficiency of these libraries in high-performance scenarios, such as those requiring rapid evolution over multiple generations or when running on resource-constrained hardware.

Most existing NEAT implementations do not utilize GPUs to accelerate computation, with a few exceptions such as PyTorchNEAT~\cite{pytorchneat}. These libraries integrate tensor-based deep learning frameworks like PyTorch~\cite{pytorch} and TensorFlow~\cite{tensorflow}, allowing networks generated by NEAT to perform inference on GPUs. This integration enables batch inference, which improves network inference speed, particularly when handling a large volume of input data. However, these libraries primarily focus on optimizing the inference aspect of the networks and still rely heavily on the object-oriented programming paradigm. While they offer some degree of performance gain in terms of network inference, the neuroevolution process—where NEAT evolves networks through mutation and crossover—remains unoptimized for GPU execution. Specifically, the computationally intensive process of evolving a population of neural networks has not been fully parallelized. Furthermore, although batch processing allows individual networks to leverage GPU capabilities, the evaluation of the entire population in NEAT, which is central to the algorithm's evolutionary process, continues to be executed sequentially. As a result, current NEAT implementations fail to fully exploit the high parallel processing potential of modern GPUs, leading to inefficiencies in both the search and evolutionary phases of the algorithm.

\section{Tensorization}

To overcome the limitations of current NEAT implementations and fully harness modern hardware for improved efficiency, we introduce a novel tensorization approach. As shown in Fig.~\ref{fig:pop_forward}, this method accelerates the NEAT algorithm by converting various network topologies and their operations into structured tensors, which are well-suited for GPU-based computations. By vectorizing functions within network operations, our approach enables parallel processing across the entire population of networks, allowing GPUs to be used more effectively. The following sections detail the specific tensorization techniques applied to network encoding and operations. \add{Table~\ref{tab:notations} presents the symbols used in this paper.}

\begin{figure}[t]
    \centering
    \includegraphics[width=0.98\columnwidth]{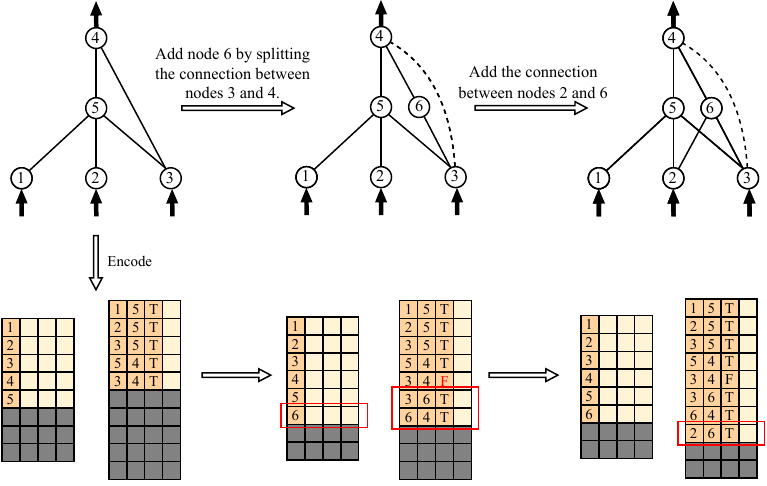}
    \caption{An illustration of tensorized network operations. It demonstrates how traditional network operations are converted into equivalent tensor operations during tensorization. Changes from the original format are highlighted in red, underscoring the modifications made within the tensor.}
    \label{fig:operations}
\end{figure}

\begin{figure*}
    \centering
    \begin{subfigure}{0.43\textwidth}
        \centering
        \includegraphics[width=\textwidth]{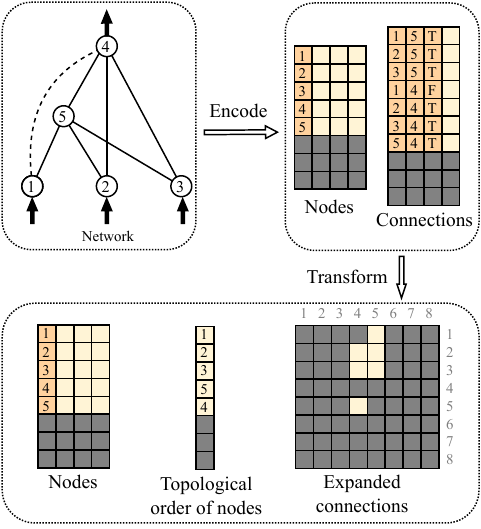}
    \end{subfigure}
    \hfill\vline\hfill
    \begin{subfigure}{0.54\textwidth}
        \centering
        \includegraphics[width=\textwidth]{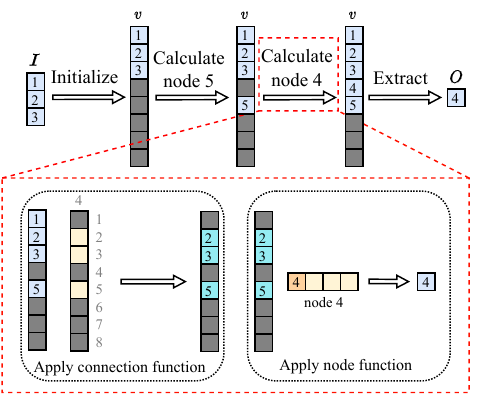}
    \end{subfigure}
    \caption{Illustration of tensorized network inference process. 
    It transforms a feedforward network's topology into tensors and the subsequent calculation of node values for network inference. 
    The network is first encoded into node and connection tensors, which are then ordered and expanded for processing. 
    Finally, values are calculated through connection and node functions to produce the output.}
    \label{fig:transform_and_forward}
\end{figure*}

\begin{table}[b]
   \renewcommand{\arraystretch}{1.2} 
    \centering
    \caption{\add{Symbols and Notations}}
    \begin{tabular}{|c|c|}
        \hline
        \textbf{Symbol} & \textbf{Explanation} \\
        \hline
        $\mathcal{N}$ & The network in the NEAT algorithm \\
        $N$ & The set of nodes in the network \\
        $C$ & The set of connections in the network \\
        $n$ & A node in the network \\
        $c$ & A connection in the network \\
        $\texttt{noa}(\cdot)$ & The number of attributes of a node or connection \\
        $\boldsymbol{n}$ & Tensor representing a single node in the network \\
        $\boldsymbol{c}$ & Tensor representing a single connection in the network \\
        $\texttt{len}(\cdot)$ & The length of a one-dimensional tensor \\
        $|\cdot|$ & The cardinality of a set \\
        $\boldsymbol{N}$ & Tensor representing all nodes in the network \\
        $\boldsymbol{C}$ & Tensor representing all connections in the network \\
        $|N|_\text{max}$ & Maximum allowed number of nodes in the network \\
        $|C|_\text{max}$ & Maximum allowed number of connections in the network \\
        $\hat{\boldsymbol{N}}$ & Nodes tensor after padding with \texttt{NaN} \\
        $\hat{\boldsymbol{C}}$ & Connections tensor after padding with \texttt{NaN} \\
        $\boldsymbol{P}_{N}$ & Tensor representing all nodes from all networks in the population \\
        $\boldsymbol{P}_{C}$ & Tensor representing all connections from all networks in the population \\
        $[]$ & Tensor indexing operation \\
        $\leftarrow$ & Tensor assignment operation \\
        \hline
    \end{tabular}
    \label{tab:notations}
\end{table}

\subsection{Tensorized Encodings}

In the NEAT algorithm, a network $\mathcal{N}$ can be represented as:
$$
\mathcal{N} = \langle N, C \rangle,
$$
where $N$, $C$ are nodes and connections in $\mathcal{N}$, respectively. They can be represented as:
$$
N = \{n_1, n_2, n_3, \ldots\} \quad \text{and} \quad C = \{c_1, c_2, c_3, \ldots\},
$$
with $n_i$ denoting the $i$-th node in $N$ and $c_i$ denoting the $i$-th connection in $C$. 

In the NEAT algorithm, each node $n$ can be represented as a tuple consisting of a historical marker and its attributes:
$$ 
n = (k, \text{attr}_1, \text{attr}_2, \ldots),
$$
where $k \in \mathbb{N} $ is the historical marking, and \(\text{attr}_i\) is the $i$-th attribute of \(n\). 
Similarly, a connection $c$ can be represented as:
\[
c = (k_i, k_o, e, \text{attr}_1, \text{attr}_2, \ldots),
\]
where $k_i \in \mathbb{N} $ and $k_o \in \mathbb{N}$ are the historical markings of the input and output nodes of $c$, respectively, $e \in \{\texttt{True}, \texttt{False}\}$ is the enabled flag. 
\add{The attributes of a node $n$ can be a bias $b$, an aggregation function $f_{\text{agg}}$ (e.g., sum), and an activation function $f_{\text{act}}$ (e.g., sigmoid), while the attributes of a connection $c$ can be a weight $w$.}

We can use one-dimensional tensors to encode $n$ and $c$:
\begin{align*}
    \boldsymbol{n} & = [k, \text{attr}_1, \text{attr}_2, \ldots] \in \mathbb{R}^{1 + \texttt{noa}(n)},\\    
    \boldsymbol{c} & = [k_i, k_o, e, \text{attr}_1, \text{attr}_2, \ldots] \in \mathbb{R}^{3 + \texttt{noa}(c)},
\end{align*}
where \(\texttt{noa}(i)\) denotes the number of attributes of \(i\).
 
Then, the sets $N$ and $C$ can be represented as tensors $\boldsymbol{N}$ and $\boldsymbol{C}$, respectively:
\begin{align*}
    \boldsymbol{N} & = [\boldsymbol{n}_1, \boldsymbol{n}_2, \ldots] \in \mathbb{R}^{|N|\times\texttt{len}(\boldsymbol{n})}, \\
    \boldsymbol{C} & = [\boldsymbol{c}_1, \boldsymbol{c}_2, \ldots] \in \mathbb{R}^{|C|\times\texttt{len}(\boldsymbol{c})},
\end{align*}
where $|\cdot|$ is the cardinality of a set, $\texttt{len}(\cdot)$ is the length of a one-dimensional tensor, and $\boldsymbol{n}_i$ and $\boldsymbol{c}_j$ are the tensor representations of the $i$-th node and $j$-th connection, respectively.   

To address the issue where each network in the population possesses a varying number of nodes and connections, we employ tensor padding using \texttt{NaN} values for alignment. We set predefined maximum limits for the number of nodes and connections, represented as $|N|_\text{max}$ and $|C|_\text{max}$, respectively. The resulting padded tensors, $\hat{\boldsymbol{N}}$ and $\hat{\boldsymbol{C}}$, are formulated as:
\begin{align*}
\hat{\boldsymbol{N}}& = [\boldsymbol{n}_1, \boldsymbol{n}_2, \ldots, \texttt{NaN}, \ldots] \in \mathbb{R}^{|N|_\text{max}\times\texttt{len}(\boldsymbol{n})}, \\
\hat{\boldsymbol{C}}& = [\boldsymbol{c}_1, \boldsymbol{c}_2, \ldots, \texttt{NaN}, \ldots] \in \mathbb{R}^{|C|_\text{max}\times\texttt{len}(\boldsymbol{c})}. 
\end{align*} 

Upon alignment, the tensors $\hat{\boldsymbol{N}}$ and $\hat{\boldsymbol{C}}$ of each network in the population can be concatenated, allowing us to express the entire population using two tensors: $\boldsymbol{P}_{N}$ and $\boldsymbol{P}_{C}$. These tensors encompass all nodes and connections within the population, respectively. Formally, the concatenated tensors, $\boldsymbol{P}_{N}$ and $\boldsymbol{P}_{C}$, are defined as:
\begin{align*}
\boldsymbol{P}_{N} & = [\hat{\boldsymbol{N}}_1, \hat{\boldsymbol{N}}_2, \ldots] \in \mathbb{R}^{P \times |N|_\text{max} \times\texttt{len}(\boldsymbol{n})}, \\
\boldsymbol{P}_{C} & = [\hat{\boldsymbol{C}}_1, \hat{\boldsymbol{C}}_2, \ldots] \in \mathbb{R}^{P \times |C|_\text{max} \times\texttt{len}(\boldsymbol{c})},
\end{align*}
where $P$ denotes the population size, $\hat{\boldsymbol{N}}_i$ and $\hat{\boldsymbol{C}}_i$ represent the tensor representations of the node and connection sets of the $i$-th network, respectively.  
Fig.~\ref{fig:pop_encoding} provides a graphical representation of the tensorized encoding process.

\subsection{Tensorized Operations}
Upon encoding networks as tensors, we subsequently express operations on NEAT networks as corresponding tensor operations. In this subsection, we detail the tensorized representations of three fundamental operations: node modification, connection modification, and attribute modification.

\subsubsection{Node Modification}
Given a network \(\mathcal{N} = \langle N, C \rangle\), node set modifications in \(N\) can involve either the addition of a new node \(n\) or the removal of an existing node \(n\):
\begin{align*}
N' = N \cup \{n\} \quad \text{or} \quad N' = N \setminus \{n\}.
\end{align*}

In the tensorized representation $\hat{\boldsymbol{N}} \in \mathbb{R}^{|N|_\text{max}\times\texttt{len}(\boldsymbol{n})} $ of the node set \(N\), the tensorized operation for node addition can be represented as:
\[
\hat{\boldsymbol{N}}[r_i] \leftarrow \boldsymbol{n}_\text{new},
\]
where \(r_i\) denotes the index of the first \texttt{NaN} row in \(\hat{\boldsymbol{N}}\), \(\boldsymbol{n}_\text{new}\) stands for the tensor representation of the node being added, \([\cdot]\) represents tensor slicing, and \(\leftarrow\) indicates the assignment operation.

Conversely, the tensorized operation for node removal can be depicted as:
\[
\hat{\boldsymbol{N}}[r_j] \leftarrow \texttt{NaN},
\]
with \(r_j\) representing the index of the node for removal.

\subsubsection{Connection Modification}
Given a network \(\mathcal{N} = \langle N, C \rangle\), modifications in the connection set \(C\) can entail either adding a new connection or eliminating an existing connection \(c\):
\begin{align*}
    C' = C \cup \{c\} \quad \text{or} \quad C' = C \setminus \{c\}.
\end{align*}

In the tensorized representation, \(\hat{\boldsymbol{C}}\), of \(C\), the operations of connection addition or removal can be depicted as:
\[
    \hat{\boldsymbol{C}}[r_i] \leftarrow \boldsymbol{c}_\text{new} \quad \text{or} \quad 
    \hat{\boldsymbol{C}}[r_j] \leftarrow \texttt{NaN}, 
\]
respectively. Here, \(r_i\) indicates the index of the initial \texttt{NaN} row in \(\hat{\boldsymbol{C}}\), \(\boldsymbol{c}_\text{new}\) represents the tensor form of the connection being introduced, and \(r_j\) signifies the index of the connection for removal.

\subsubsection{Attribute Modification}
NEAT is designed not only to modify the network structures but also the internal attributes, either in a node or a connection. Specifically, given a node $n = (k, \text{attr}_1, \text{attr}_2, \ldots)$, 
where $k \in \mathbb{N} $ is the historical marking, and \(\text{attr}_i\) is the $i$-th attribute of $n$, 
when modifying the \(j\)-th attribute in a node, the transformation can be represented as:
\[
n' = (k, \text{attr}_1, \text{attr}_2, \ldots, \text{attr}'_j, \ldots),
\]
and the corresponding tensorized operation is:
\[
\boldsymbol{n}[1 + j] \leftarrow \text{attr}'_j.
\]
Similarly, for a connection $c = (k_i, k_o, e, \text{attr}_1, \text{attr}_2, \ldots)$, where $k_i \in \mathbb{N} $ and $k_o \in \mathbb{N}$ are the historical markings of the input and output nodes of $c$, respectively, $e \in \{\text{True}, \text{False}\}$ is the enabled flag, when modifying the \(j\)-th attribute in a connection, the transformation can represented as:
\[
c' = (k_i, k_o, e, \text{attr}_1, \text{attr}_2, \ldots, \text{attr}'_j, \ldots),
\]
and the corresponding tensorized operation is:
\[
\boldsymbol{c}[3 + j] \leftarrow \text{attr}'_j.
\]

By combining the aforementioned three operations, operations for searching networks in the NEAT algorithm including Mutation and Crossover can be transformed into operations on tensors. Fig.~\ref{fig:operations} provides a graphical representation of the tensorized network operations.

\subsection{Tensorized Network Inference}
Another crucial component in NEAT is the inference process, where a network receives inputs and generates corresponding outputs based on its topologies and weights. For a network $\mathcal{N}$, given its node tensor $\hat{\boldsymbol{N}}$ and connection tensor $\hat{\boldsymbol{C}}$, the inference process can be represented as:
$$
\bm{O} = \text{inference}_{\hat{\boldsymbol{N}}, \hat{\boldsymbol{C}}}(\bm{I}), 
$$
where $\bm{O}$ and $\bm{I}$ denote the outputs and inputs in the inference process, respectively.

In our tensorization method, the inference process is divided into two stages: transformation and calculation. `Transformation' involves converting the node tensor $\hat{\boldsymbol{N}}$ and connection tensor $\hat{\boldsymbol{C}}$ into formats more conducive to network inference. 
`Calculation' refers to computing the output using the tensors produced in the transformation stage.
When a network undergoes multiple inference operations, it only needs to be transformed once. Networks in the NEAT algorithm can be categorized as either feedforward or concurrent, based on the presence or absence of cycles in their topological structure. Here, we primarily focus on the transformation and calculation processes in feedforward networks.

In feedforward networks, the transformation process creates two new tensors: the topological order of nodes $\boldsymbol{N}_{\text{order}} \in \mathbb{R}^{|N|_\text{max}}$ and the expanded connections $\boldsymbol{C}_{\text{exp}} \in \mathbb{R}^{|N|_\text{max} \times |N|_\text{max} \times \texttt{noa}(c)}$, where $|N|_\text{max}$ represents the predefined maximum limit for the number of nodes and $\texttt{noa}(c)$ denoting the number of attributes of connections in the network.

\add{Given the absence of cycles in the network, topological sorting~\cite{topological_sort} is employed to obtain $\boldsymbol{N}_{\text{order}}$.} For $\boldsymbol{C}_{\text{exp}}$, the generation rule can be represented as:
$$
    \boldsymbol{C}_{\text{exp}}[\hat{\boldsymbol{C}}[i][0, 1]] \leftarrow 
    \begin{cases}
        \texttt{NaN}, & \hat{\boldsymbol{C}}[i][2] = 0\\
        \hat{\boldsymbol{C}}[i][2:], & \hat{\boldsymbol{C}}[i][2] = 1
    \end{cases},
$$
where $i = 0, 1, 2, \ldots, |C|_\text{max}$, and $|C|_\text{max}$ is the predefined maximum limit for the number of connections. 
Recall that in each line $c$ of $\hat{\boldsymbol{C}}$, the values are $(k_i, k_o, e, \text{attr}1, \text{attr}2, \ldots)$, with $c[0]=k_i$ and $c[1]=k_o$ indicating the indices of the input and output nodes of the connection, respectively, and $c[2]=e \in \{\texttt{True}, \texttt{False}\}$ representing the enabled flag.
Locations not updated by this rule default to the value $\texttt{NaN}$. The tensors $\hat{\boldsymbol{N}}$, $\boldsymbol{N}_{\text{order}}$, and $\boldsymbol{C}_{\text{exp}}$ are then used as inputs for the calculation process.

In the forward process, we utilize the transformed tensors $\hat{\boldsymbol{N}}$, $\boldsymbol{N}_{\text{order}}$, $\boldsymbol{C}_{\text{exp}}$, and the input $\bm{I}$ to calculate the output $\bm{O}$. We maintain a tensor $\boldsymbol{v} \in \mathbb{R}^{|N|_\text{max}}$ to store the values of nodes in the network. Initially, $\boldsymbol{v}$ is set to the default value $\texttt{NaN}$ and then updated with:
$$
\boldsymbol{v}[k_{\text{input}}] \leftarrow \bm{I},
$$
where $k_{\text{input}}$ denotes the indices of input nodes in the network.
The value of nodes is iteratively calculated in the order specified by $\boldsymbol{N}_{\text{order}}$.
The rule to obtain the value $\boldsymbol{v}[k]$ of node $\boldsymbol{n}_k$ can be expressed as:
\begin{align*}
\boldsymbol{v}[k] \leftarrow f_n(f_c(\boldsymbol{v} \ |\  \boldsymbol{C}_{\text{exp}}[:][k]) \ | \ \hat{\boldsymbol{N}}[k][1:]),
\end{align*}
where $\boldsymbol{C}_{\text{exp}}[:][k]$ indicates the attributes of all connections to $\boldsymbol{n}_k$, and $\hat{\boldsymbol{N}}[k][1:]$ denotes the attributes of $\boldsymbol{n}_k$. $f_c$ and $f_n$ are the calculation functions for connections and nodes in the network, respectively. For a network with connection attribute weight $w$, and node attributes bias $b$, aggregation function $f_{\text{agg}}$ and activation function $f_{\text{act}}$, $f_c$ and $f_n$ can be defined as:
\begin{align*}
    f_c(\bm{I}_{\text{conn}}\ |\ w) &= w\bm{I}_{\text{conn}},\\
    f_n(\bm{I}_{\text{node}}\ |\ b, f_{\text{agg}}, f_{\text{act}}) &= f_{\text{act}}(f_{\text{agg}}(\bm{I}_{\text{node}}) + b),
\end{align*}
where $\bm{I}_{\text{conn}}$ and $\bm{I}_{\text{nodes}}$ denote the inputs for connections and nodes, respectively.

After computing the values of all nodes, the tensor $\boldsymbol{v}[k_{\text{output}}]$ represents the network's output, with $k_{\text{output}}$ indicating the indices of output nodes in the network.
Fig.~\ref{fig:transform_and_forward} graphically depicts the tensorized network inference process.

\add{Tensorization fundamentally embodies a space-time trade-off. By setting upper limits on network size, operations can be executed with fixed-shaped tensors on GPUs, maximizing parallel computation efficiency. Setting appropriate $|N|_\text{max}$ and $|C|_\text{max}$ is crucial as values that are too low restrict network complexity growth, while excessively high values waste GPU memory and reduce efficiency. In practice, these limits should be chosen based on problem requirements, and iterative tuning is often necessary to balance network scalability and computational speed.}

\section{Implementation}

In this section, we outline the implementation of TensorNEAT, a library designed to enhance NEAT's scalability through tensorization and GPU/TPU acceleration via JAX. By converting network topologies into uniform tensors, TensorNEAT enables efficient parallelization of evolutionary processes, significantly boosting performance. We discuss its key components, including JAX-based hardware acceleration, user-friendly customization interfaces, and intuitive network visualization tools.

\subsection{JAX-based Hardware Acceleration}
JAX \cite{frostig2018compiling} is an open-source numerical computing library, offering APIs similar to NumPy \cite{harris2020array} and enabling efficient execution across various hardware platforms (CPU/GPU/TPU). Leveraging the capabilities of XLA, JAX facilitates the transformation of numerical code into optimized machine instructions. The optimization techniques provided by JAX have supported the development of numerous projects in evolutionary computation, as evidenced by \cite{tang2022evojax}, \cite{lange2022evosax}, \cite{lim2022qdax}, and \cite{huang2023evox}. These advancements significantly contribute to JAX's growing popularity in the scientific community. Furthermore, JAX's flexibility and ease of integration with other machine learning frameworks, such as TensorFlow and PyTorch, have enabled researchers to experiment with cutting-edge models while benefiting from GPU acceleration and automatic differentiation. This has positioned JAX as a preferred tool for both traditional deep learning research and more specialized domains, like neuroevolution and reinforcement learning, where computational efficiency and scalability are crucial. As a result, JAX continues to be a key enabler for new methods and approaches in the field of evolutionary algorithms, pushing the boundaries of what can be achieved in terms of speed and performance.

TensorNEAT, integrating JAX, utilizes the functional programming paradigm to implement our proposed tensorization methods. This integration allows NEAT to effectively use hardware accelerators such as GPUs and TPUs, enabling significant speedups in evolutionary processes. By maintaining uniform tensor shapes in network encoding, several key NEAT operations, such as mutation, crossover, and network inference, can be efficiently vectorized across the entire population dimension. This approach not only simplifies the computation but also ensures scalability, as operations can be parallelized and executed on large populations without compromising performance. Specifically, the use of JAX's \texttt{jax.vmap} function enables automatic vectorization of operations over multiple individuals, while \texttt{jax.pmap} allows for further parallelization across multiple devices in a multi-GPU or multi-TPU setup. These combined capabilities make TensorNEAT adaptable to both small and large-scale experiments, facilitating faster training times and broader exploration of the search space, especially when dealing with complex neural architectures. Additionally, TensorNEAT's design ensures that these enhancements are implemented without altering the core principles of NEAT, preserving its evolutionary behavior.

\subsection{User-friendly Interfaces}
Designed with user-friendly interfaces, TensorNEAT provides mechanisms for adjusting algorithms to specific requirements. It offers an extensive set of hyperparameters, allowing users to fine-tune various computational elements of the NEAT algorithm. Additionally, its modular problem templates facilitate the integration of specialized problems, ensuring adaptability to diverse application domains. By implementing a few functions, users can define the behavior of networks within the NEAT algorithm, making the customization process both flexible and straightforward. A notable feature is the interface supporting the evolution of advanced network architectures, including Spiking Neural Networks \cite{ghosh2009spiking} and Binary Neural Networks \cite{hubara2016binarized}, which significantly broadens the scope of potential experiments. Furthermore, the system is designed to handle complex tasks with minimal setup, enabling users to quickly prototype new architectures or modify existing ones. This ease of use, coupled with its comprehensive support for cutting-edge network types, makes TensorNEAT a versatile tool for neuroevolution research. Details on the hyperparameters and the interfaces are elaborated in Appendix~\ref{Appendix_a} and Appendix~\ref{Appendix_b}, ensuring users have thorough guidance for fine-tuning and customization.

\subsection{Intuitive Network Visualization}
\begin{figure*}
    \centering
    \begin{subfigure}{0.43\textwidth}
        \centering
        \includegraphics[width=\textwidth]{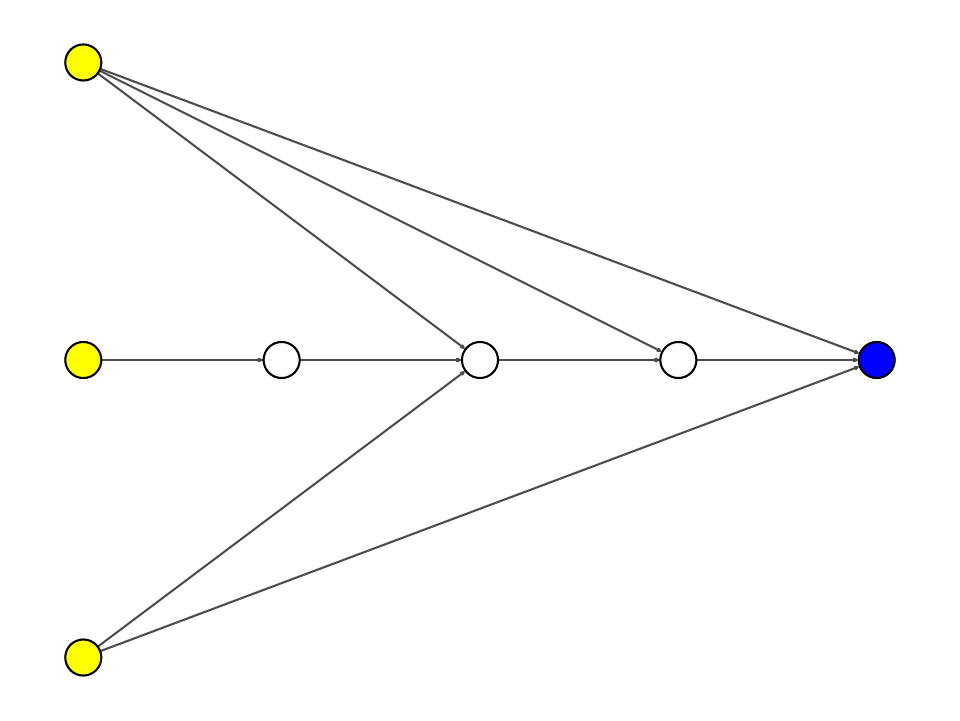}
    \end{subfigure}
    \hfill\vline\hfill
    \begin{subfigure}{0.54\textwidth}
        \centering
        \includegraphics[width=\textwidth]{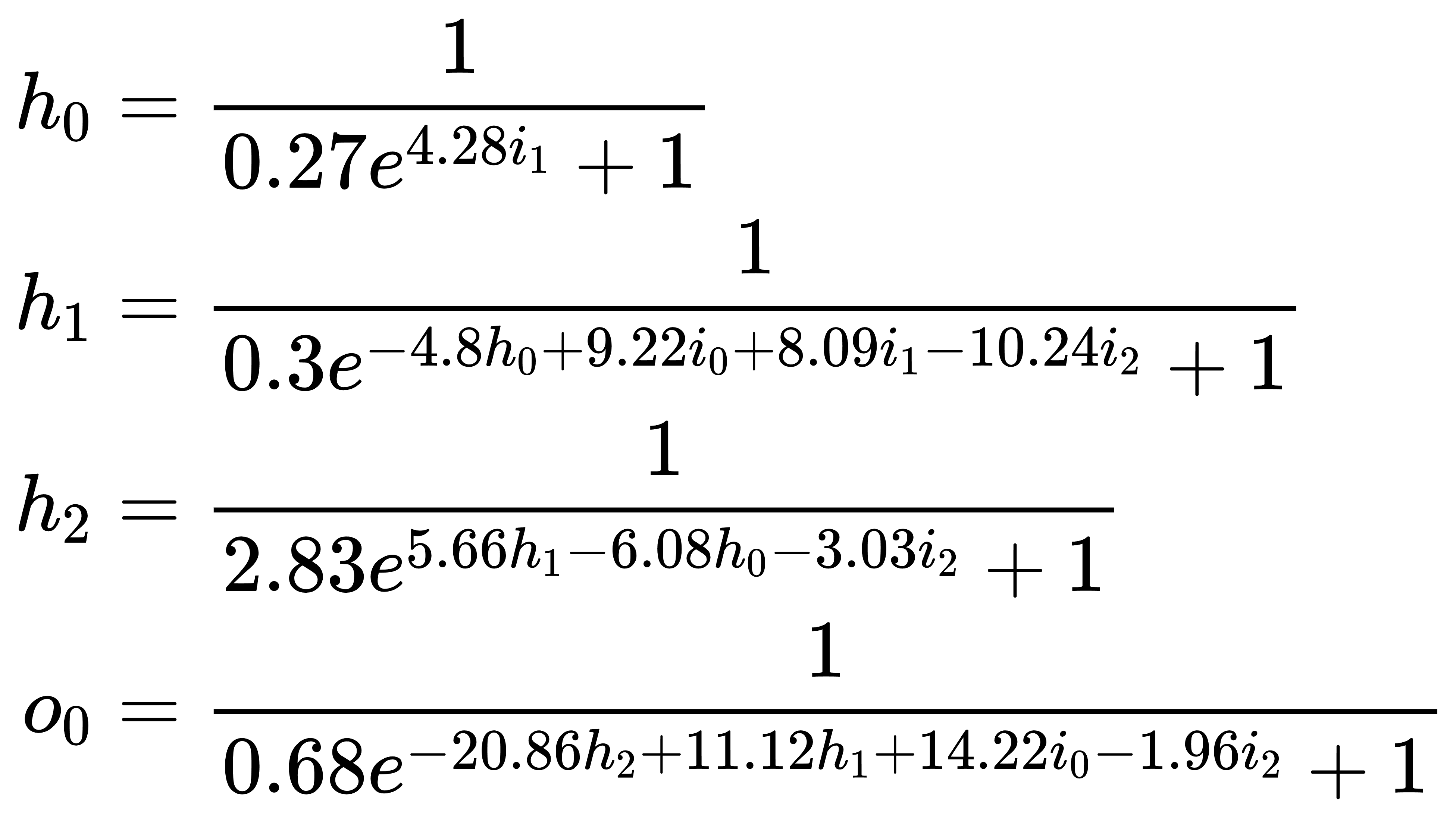}
    \end{subfigure}
    \caption{Visualization methods for networks in TensorNEAT. The left side shows the topology diagram, and the right side represents the network formulations. This network has three input nodes, one output node, and three hidden nodes. In the topology diagram, yellow nodes represent input nodes, white nodes represent hidden nodes, and the blue node represents the output node. In the formula, the values of the input nodes are denoted by \(i_0\), \(i_1\), and \(i_2\); the hidden nodes by \(h_0\), \(h_1\), and \(h_2\); and the output node by \(o_0\).}
    \label{fig:visualization}
\end{figure*}

TensorNEAT provides two ways to visualize networks: topology diagrams and network formulations. The topology diagram visually represents the structure of the network by displaying the arrangement of nodes and connections, making it easier to understand the architecture and flow of information within the network. On the other hand, network formulations allow users to convert NEAT-generated networks into more formal representations, such as Latex formulas and Python code. This feature is particularly useful for researchers who want to mathematically analyze the network or integrate it into other software environments. These visualizations offer both an intuitive and a formal way of interacting with the evolved networks, enhancing both usability and interpretability. Fig.~\ref{fig:visualization} demonstrate these two visualization methods. Together, they provide a comprehensive view of the networks, from their graphical structure to their formalized representations, enabling better understanding and facilitating further experimentation or integration into different workflows.

\subsection{Feature-rich Extensions}
Extending beyond the conventional NEAT paradigm, TensorNEAT includes notable algorithmic extensions such as Compositional Pattern Producing Networks (CPPN) \cite{stanley2007compositional} and HyperNEAT \cite{stanley2009hypercube}, tailored for parallel processing on hardware accelerators. In addition to these algorithmic improvements, TensorNEAT is designed to leverage the computational capabilities of modern GPUs, enabling faster and more scalable evolution processes compared to traditional CPU-based approaches. For evaluation, TensorNEAT provides a comprehensive suite of standard test benchmarks, which span a wide range of domains from numerical optimization to complex function approximation tasks, ensuring broad applicability across various problem types. Moreover, TensorNEAT integrates seamlessly with leading reinforcement learning environments like Gym \cite{brockman2016openai}, and hardware-optimized platforms such as gymnax \cite{lange2022gymnax} and Brax \cite{freeman2021brax}, allowing users to evaluate the performance of the NEAT algorithm in both simulated and hardware-accelerated settings. This flexibility makes TensorNEAT a robust and versatile tool for researchers and practitioners looking to apply NeuroEvolution techniques in high-performance computing environments.

\section{Experiments}

This section presents a comprehensive evaluation of TensorNEAT, highlighting its performance across diverse robotic control tasks and hardware configurations. The experiments are designed to validate the effectiveness of the proposed tensorization method in addressing the computational challenges inherent in the NEAT algorithm. Specifically, we aim to answer the following key questions:
\begin{enumerate}
    \item How does TensorNEAT compare with the existing NEAT implementation (NEAT-Python) in terms of execution time and solution quality?
    \item How does TensorNEAT perform on multi-GPU setups?
    \item What is the practical impact of the parallelism introduced by tensorization on the NEAT algorithm?
    \item How does TensorNEAT compare with other GPU-accelerated algorithms in terms of execution time and solution quality?
\end{enumerate}
By systematically addressing these questions, we demonstrate the scalability and practical benefits of the proposed approach.

\subsection{Experimental Setup}
This section outlines the experimental configuration used to evaluate TensorNEAT. The setup is designed to assess the library's performance across a variety of tasks, hardware configurations, and parameter settings.

\begin{figure}[htbp]
    \centering
    \begin{subfigure}{0.23\columnwidth}
        \centering
        \includegraphics[width=\columnwidth]{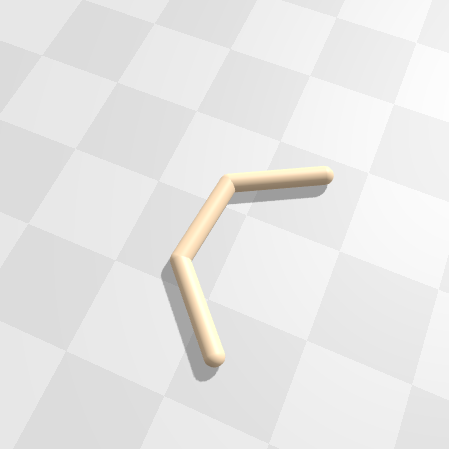}
        \caption{Swimmer}
        \label{fig:swimmer}
    \end{subfigure}
    \begin{subfigure}{0.23\columnwidth}
        \centering
        \includegraphics[width=\columnwidth]{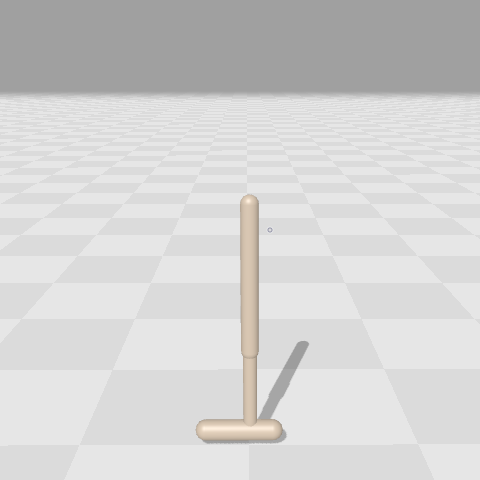}
        \caption{Hopper}
        \label{fig:hopper}
    \end{subfigure}
    \begin{subfigure}{0.23\columnwidth}
        \centering
        \includegraphics[width=\columnwidth]{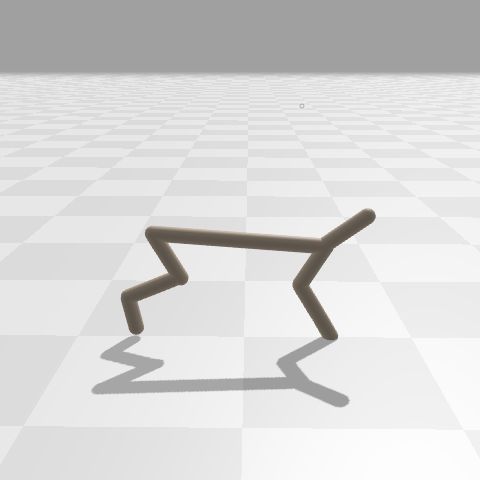}
        \caption{Halfcheetah}
        \label{fig:halfcheetah}
    \end{subfigure}
    \caption{Robotic control tasks in Brax.}
    \label{fig:tasks}
\end{figure}

Three tasks in Brax~\cite{freeman2021brax} including Swimmer, Hopper and HalfCheetah are used to test TensorNEAT. These robotic control tasks require the robots to advance as far as possible in a forward direction. Fig.~\ref{fig:tasks} displays representative screenshots of these environments, while Table~\ref{table:experiment settings} summarizes their key attributes.

\begin{table}[htbp]
    \centering
    \begin{tabular}{|c|c|c|c|c|}
        \hline
        Environment & Action Dimension & Observation Dimension & Termination Check \\
        \hline
        Swimmer & 2 & 8 & False \\
        \hline
        Hopper & 3 & 11 & True \\
        \hline
        Halfcheetah & 6 & 18 & False \\
        \hline
    \end{tabular}
    \caption{Attributes of the Brax environments}
    \label{table:experiment settings}
\end{table}

Details of the hardware specifications are provided in Table~\ref{tab:system_specifications}. In the experiments, unless otherwise specified, we used the Nvidia RTX 4090 GPU. To evaluate the performance of the algorithms, we varied parameters such as population size and the maximum number of generations while keeping other parameters fixed. Some of the key parameters used by TensorNEAT are listed in Table~\ref{tab:params}. For experiments involving other libraries including NEAT-Python~\cite{McIntyre_neat-python} and evosax~\cite{lange2022evosax}, we employed their default parameter settings. A complete description of the experimental parameters is provided in Appendix~\ref{Appendix_c}.

\begin{table}[h!]
    \centering
    \begin{minipage}{0.45\textwidth}
        \centering
        \caption{Hardware Specifications}
        \begin{tabular}{|c|c|}
            \hline
            \textbf{Component} & \textbf{Specification} \\
            \hline
            CPU & AMD EPYC 7543 \\
            CPU Cores & 8 \\
            GPU (Default) & NVIDIA RTX 4090 \\
            Host RAM & 512 GB \\
            GPU RAM & 24 GB \\
            OS & Ubuntu 22.04.4 LTS \\
            \hline
        \end{tabular}
        \label{tab:system_specifications}
    \end{minipage}%
    \begin{minipage}{0.45\textwidth}
        \centering
        \caption{Core Parameters of TensorNEAT}
        \begin{tabular}{|c|c|}
            \hline
            \textbf{Parameter} & \textbf{Value} \\
            \hline
            max\_nodes & $50$ \\
            max\_conns & $100$ \\
            max\_species & $10$ \\
            node\_add\_prob & $0.2$ \\
            conn\_add\_prob & $0.4$ \\
            survival\_threshold & $0.2$ \\
            \hline
        \end{tabular}
        \label{tab:params}
    \end{minipage}
\end{table}



All experiments were repeated 10 times using different random seeds to ensure statistical robustness. Results are reported as mean values with 95\% confidence intervals, providing a reliable basis for comparison.

\subsection{Comparison with NEAT-Python}
\begin{figure}[t]
    \centering
    \begin{subfigure}{0.32\textwidth}
        \centering
        \includegraphics[width=\textwidth]{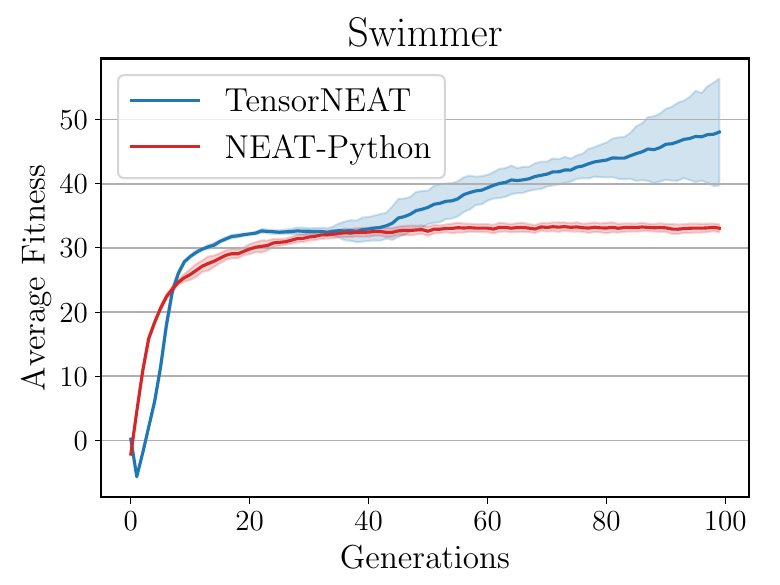}
    \end{subfigure}
    \hfill
    \begin{subfigure}{0.32\textwidth}
        \centering
        \includegraphics[width=\textwidth]{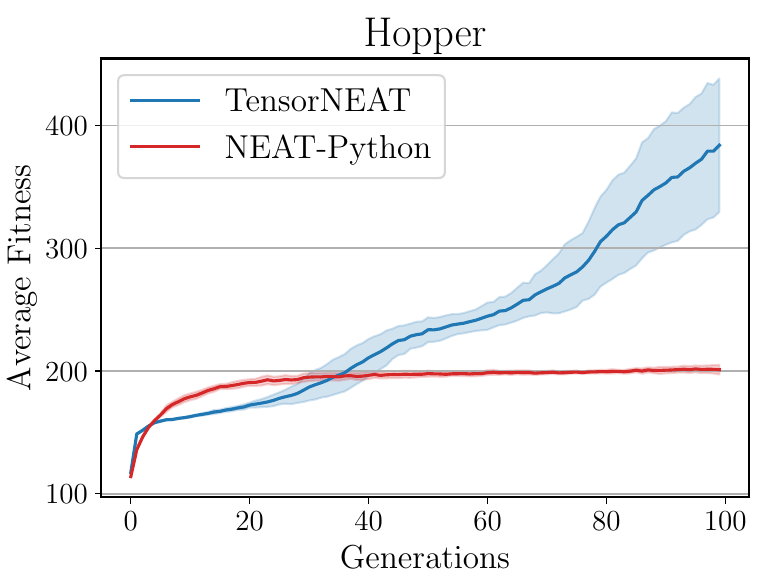}
    \end{subfigure}
    \hfill
    \begin{subfigure}{0.32\textwidth}
        \centering
        \includegraphics[width=\textwidth]{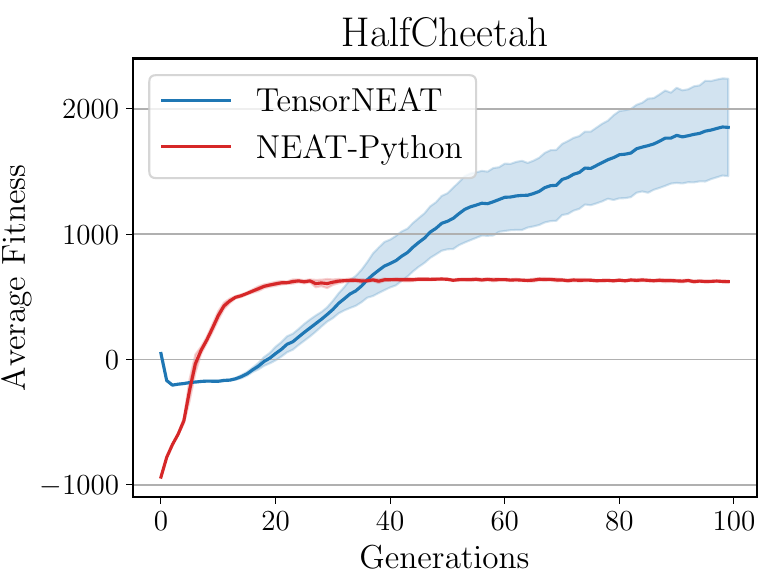}
    \end{subfigure}
    \hfill
    \begin{subfigure}{0.32\textwidth}
        \centering
        \includegraphics[width=\textwidth]{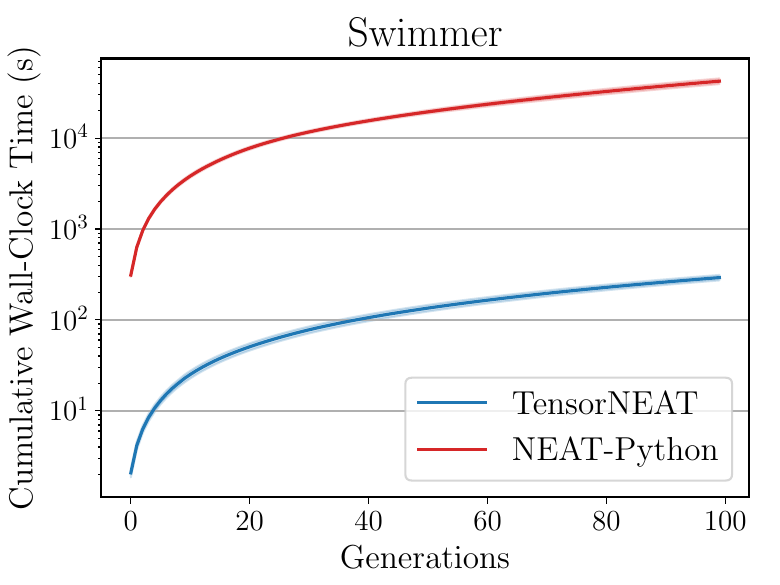}
    \end{subfigure}
    \hfill
    \begin{subfigure}{0.32\textwidth}
        \centering
        \includegraphics[width=\textwidth]{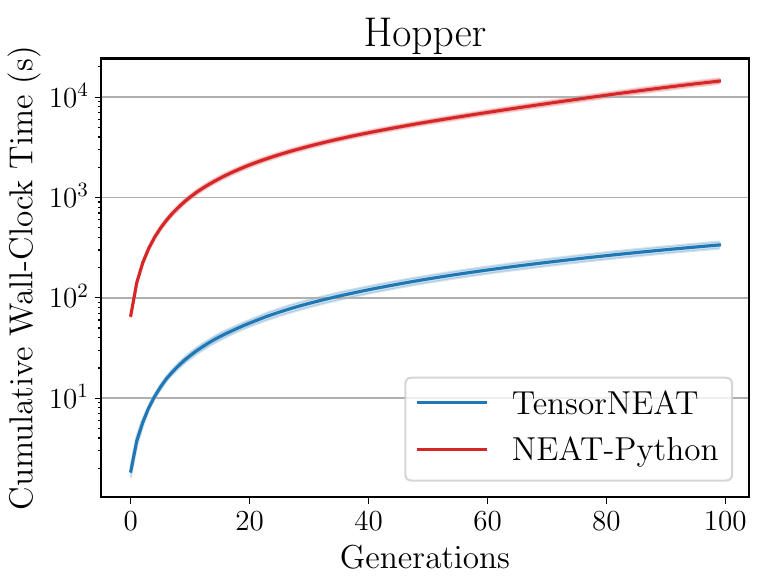}
    \end{subfigure}
    \hfill
    \begin{subfigure}{0.32\textwidth}
        \centering
        \includegraphics[width=\textwidth]{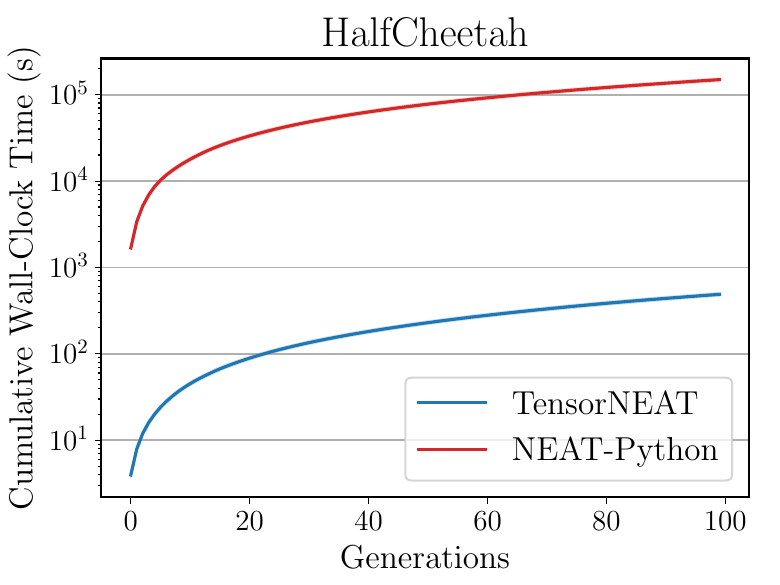}
    \end{subfigure}
        \begin{subfigure}{0.32\textwidth}
        \centering
        \includegraphics[width=\textwidth]{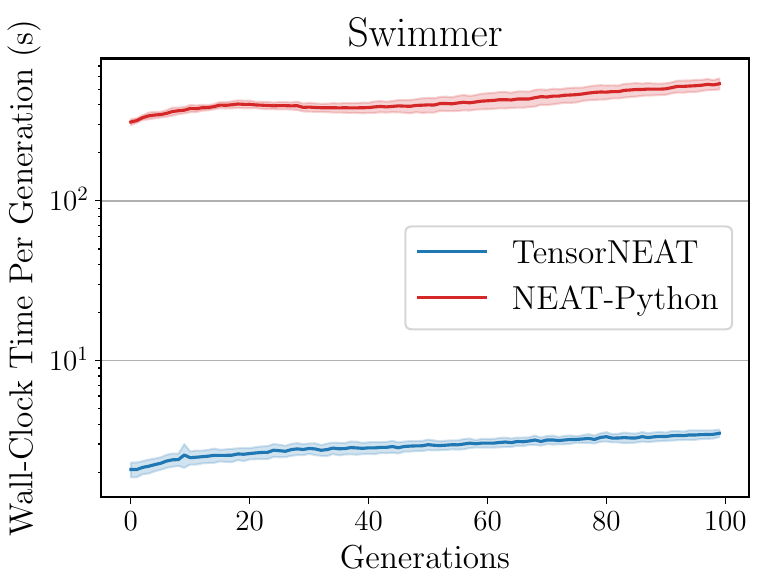}
    \end{subfigure}
    \hfill
    \begin{subfigure}{0.32\textwidth}
        \centering
        \includegraphics[width=\textwidth]{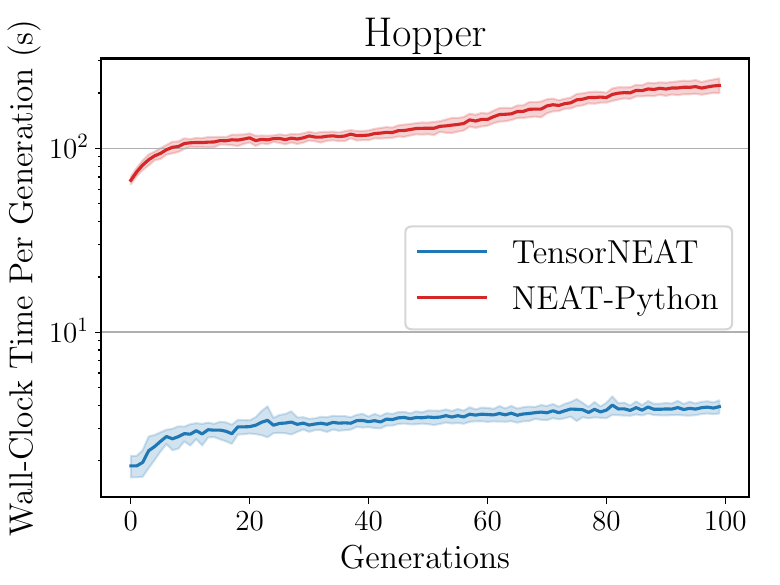}
    \end{subfigure}
    \hfill
    \begin{subfigure}{0.32\textwidth}
        \centering
        \includegraphics[width=\textwidth]{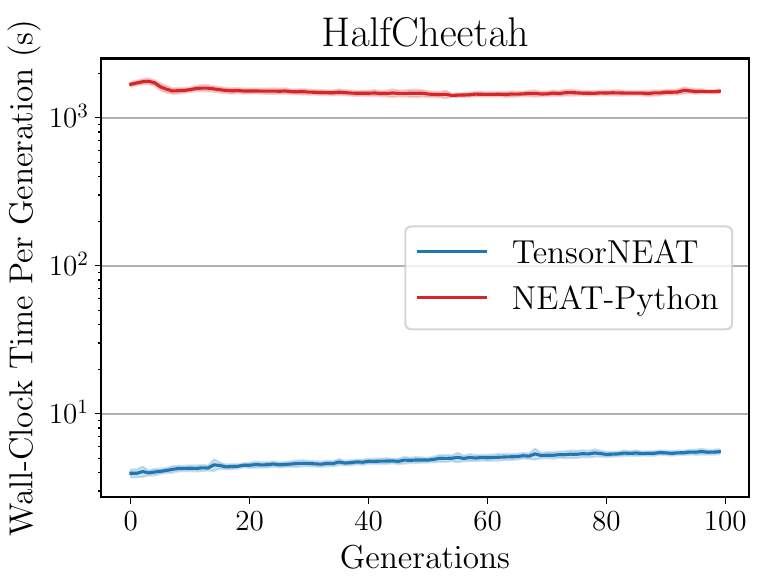}
    \end{subfigure}
    \caption{\add{Comparison of TensorNEAT and NEAT-Python with a fixed population size of $10,000$, showing population fitness, cumulative runtime, and per-generation runtime.}}

    \label{fig:popsize-constant-10000}
\end{figure}

\add{In this experiment, we compare the performance of TensorNEAT with the NEAT-Python library.}
First, we examined the evolution of average population fitness and the cumulative runtime of the algorithms across generations, with a constant population size of 10,000. 
As depicted in \add{the first row in} Fig.~\ref{fig:popsize-constant-10000}, TensorNEAT exhibits a more rapid improvement in population fitness over the course of the algorithm's execution. 
\add{We believe that the performance disparity between the two frameworks stems from modifications introduced in the NEAT algorithm after tensorization. To facilitate tensorization, TensorNEAT differs from NEAT-Python in several key aspects, such as population initialization, network distance computation, and species updating mechanisms. These differences have also resulted in variations in hyperparameter settings between TensorNEAT and NEAT-Python. We consider these discrepancies to be natural, as the purpose of this experiment is to validate that TensorNEAT provides a faithful implementation of the NEAT algorithm.}
Furthermore, the analysis of execution times, illustrated in the \add{second row} of Fig.~\ref{fig:popsize-constant-10000}, reveals a significant decrease in runtime with TensorNEAT compared to NEAT-Python.

Given the iterative nature of NEAT's process, there is an expected increase in per-iteration time as network structures become more complex. 
This phenomenon was investigated by comparing the per-generation runtimes of both algorithms. 
In the Swimmer and Hopper tasks, as shown in the \add{last row} of Fig.~\ref{fig:popsize-constant-10000}, NEAT-Python exhibits a more marked increase in runtime, likely due to its less efficient object encoding mechanism, which becomes increasingly cumbersome with rising network complexity. 
By contrast, TensorNEAT, employing a tensorized encoding approach constrained by the pre-set maximum values of $|N|_\text{max}$ and $|C|_\text{max}$, achieves a consistent network encoding size, resulting in more stable iteration times. 

\begin{figure}[t]
    \centering    
    \begin{subfigure}{0.32\textwidth}
        \centering
        \includegraphics[width=\textwidth]{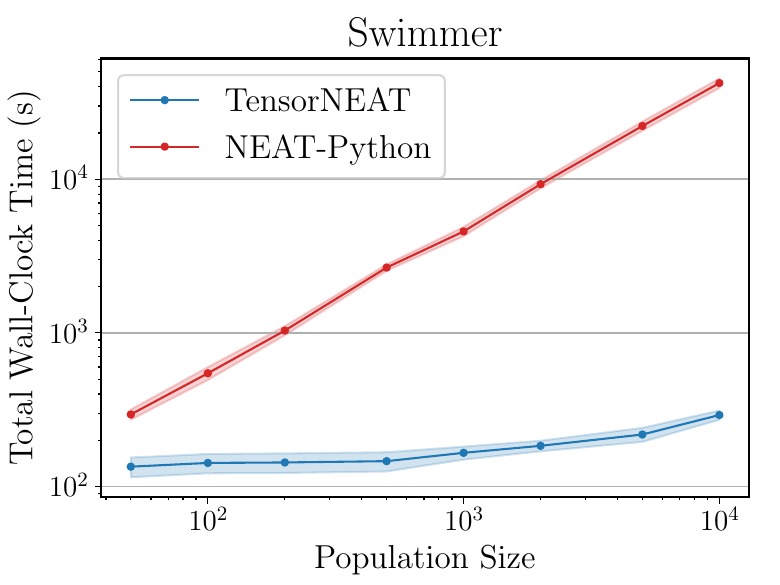}
    \end{subfigure}
    \hfill
    \begin{subfigure}{0.32\textwidth}
        \centering
        \includegraphics[width=\textwidth]{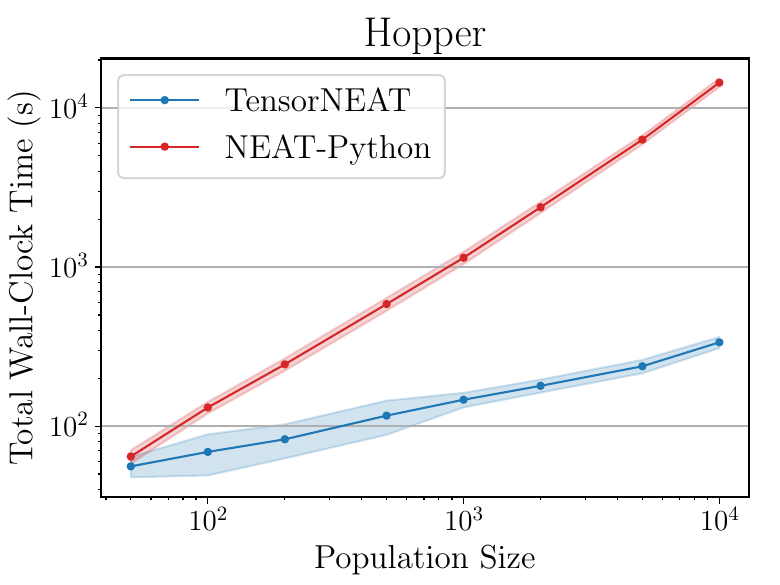}
    \end{subfigure}
    \hfill
    \begin{subfigure}{0.32\textwidth}
        \centering
        \includegraphics[width=\textwidth]{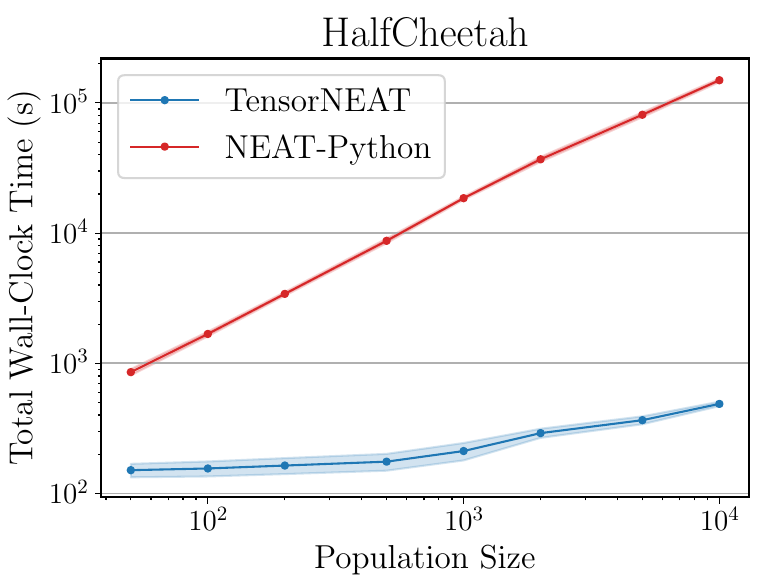}
    \end{subfigure}
    
    \caption{\add{Runtime comparison of TensorNEAT and NEAT-Python across different population sizes.}}

    \label{fig:time-generation}
\end{figure}

Additionally, we investigated how runtime varies with changes in population size, ranging from 50 to 10,000. 
As illustrated in Fig.~\ref{fig:time-generation}, NEAT-Python's runtime significantly increases with larger populations, whereas TensorNEAT only experiences a marginal increase in runtime. 
It highlights TensorNEAT's effective utilization of GPU parallel processing capabilities, further emphasizing its enhanced performance in large-scale computational tasks. 

The adaptability of TensorNEAT was further validated by evaluating its performance across various hardware configurations, including the AMD EPYC 7543 CPU and a selection of mainstream GPU models. 
These evaluations were conducted with a consistent population size of $10,000$. 
The total wall-clock time was recorded over $100$ generations of the algorithm, and the results are summarized in Table~\ref{table:time_hardware}. 
In every GPU configuration, TensorNEAT demonstrated a significant performance advantage over NEAT-Python. Notably, with the RTX 4090, TensorNEAT achieved a speedup surpassing $500\times$ in the Halfcheetah environment.
It is also noteworthy that TensorNEAT realized a speedup on the same CPU device in comparison to NEAT-Python, especially in the more complex Halfcheetah environment. 

\add{Table~\ref{table:time_hardware} shows that TensorNEAT achieves varying speedups across different Brax environments, with the highest acceleration in Halfcheetah and a lower effect in Hopper. This discrepancy arises from two factors: (1) network complexity and (2) the number of network inferences per episode. Halfcheetah has the largest observation and action spaces, leading to more complex networks that benefit significantly from TensorNEAT's GPU acceleration. In contrast, Hopper includes an early termination mechanism where poorly performing policies result in shorter episodes, while better policies require more inferences per episode. As shown in Fig.~\ref{fig:popsize-constant-10000}, TensorNEAT evolves higher-fitness policies in Hopper, increasing the overall computational workload and reducing the relative speedup. This suggests that acceleration gains depend on task characteristics, particularly network complexity and inference demands.}

\begin{table*}[htbp]
    \centering
        \caption{Runtimes over $100$ generations on different hardware configurations, with a constant population size of $10,000$.}
    \begin{tabular}{c|c|c|c|c}
            
        \hline
        Task & Library & Hardware & Time ($s$) &Speedup\\
        \hline
        
        \multirow{5}{*}{Swimmer} & NEAT-Python & EPYC 7543 (CPU) & $42279.63 \pm 3031.97$ & $1.00$ \\
        
        \cline{2-5}
        & \multirow{4}{*}{TensorNEAT} & RTX 4090 & $\bm{215.70 \pm 6.35}$ & $\bm{196.01}$\\  
                                    && RTX 3090 & $292.22 \pm 20.19$ & $144.68$\\
                                    && RTX 2080Ti & $434.94 \pm 12.49$ & $97.21$\\
                                    && EPYC 7543 (CPU) & $14678.10 \pm 616.85$ & $2.88$\\

        \hline
        \multirow{5}{*}{Hopper} & NEAT-Python & EPYC 7543 (CPU) & $14438.13 \pm 900.68$ & $1.00$\\
        
        \cline{2-5}
        & \multirow{4}{*}{TensorNEAT} & RTX 4090 & $\bm{241.30 \pm 9.41}$ & $\bm{59.83}$\\  
                                    && RTX 3090 & $336.08 \pm 26.96$ & $42.96$\\
                                    && RTX 2080Ti & $518.40 \pm 7.22$ & $27.85$\\
                                    && EPYC 7543 (CPU) & $13473.01 \pm 544.07$ & $1.07$\\
                                    
        \hline
        
        \multirow{5}{*}{Halfcheetah} & NEAT-Python & EPYC 7543 (CPU) & $149516.00 \pm 4817.90$ & $1.00$\\
        \cline{2-5}
        & \multirow{4}{*}{TensorNEAT} & RTX 4090 & $\bm{274.74 \pm 14.21}$ &  $\bm{544.21}$\\  
                                    && RTX 3090 & $487.82 \pm 19.05$ & $306.50$\\
                                    && RTX 2080Ti & $705.13 \pm 16.02$ &  $212.04$\\
                                    && EPYC 7543 (CPU) & $15914.08 \pm 4005.08$ & $9.40$\\
        \hline
    \end{tabular}
    \label{table:time_hardware}
\end{table*}

\subsection{Performance on Multiple GPUs}
\begin{figure}[t]
\centering
    \begin{subfigure}{0.48 \textwidth}
        \centering
        \includegraphics[width=\textwidth]{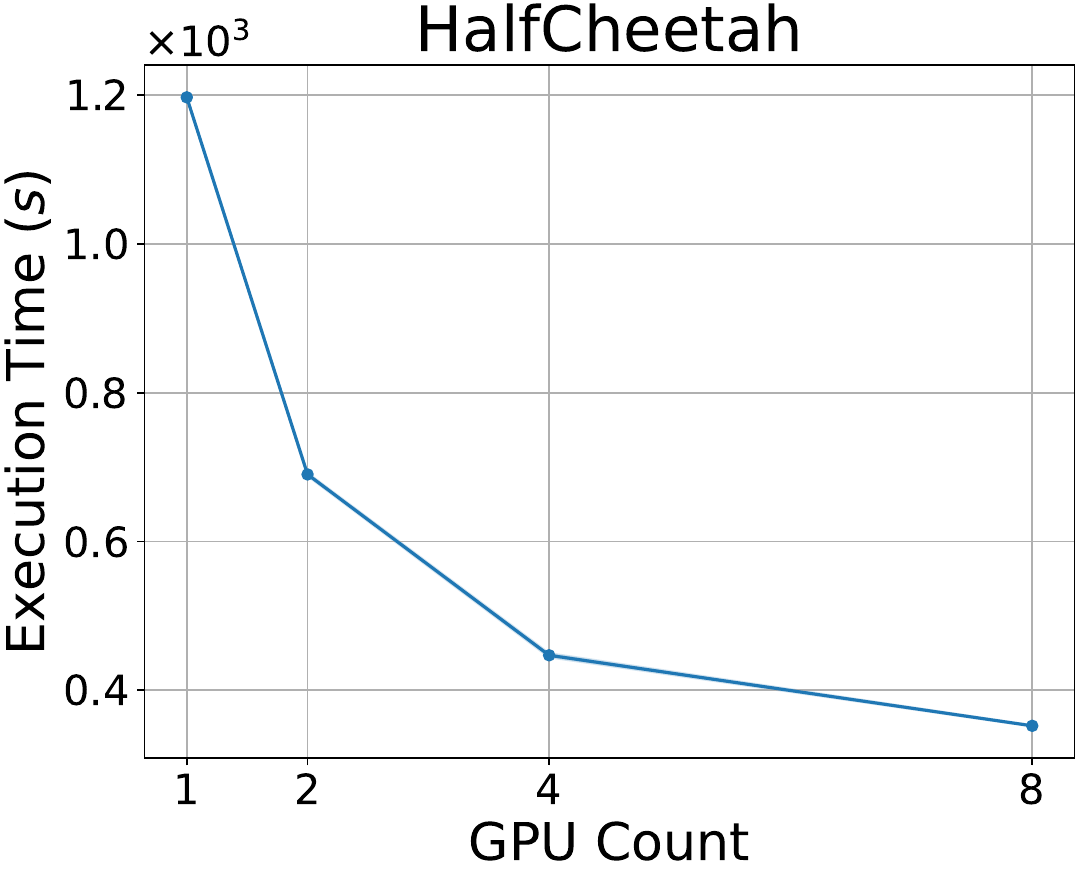}
    \end{subfigure}
    \hfill
    \begin{subfigure}{0.48\textwidth}
        \centering
        \includegraphics[width=\textwidth]{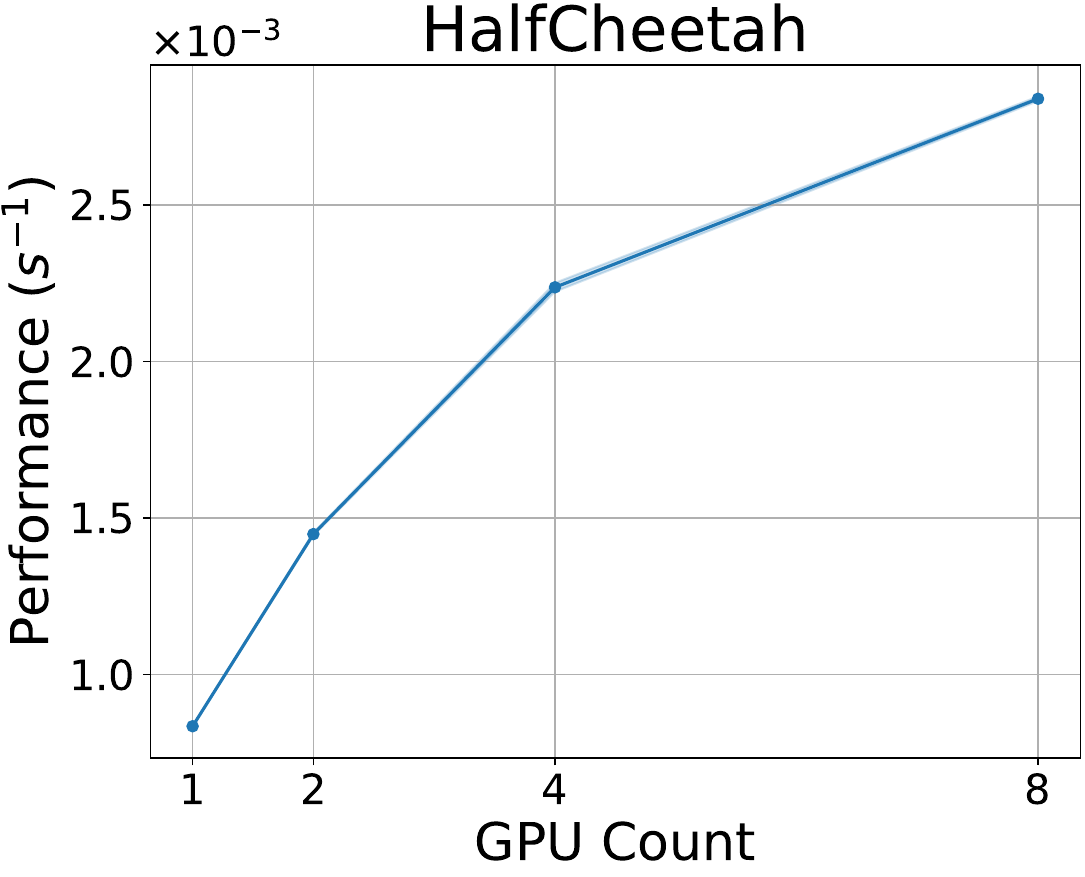}
    \end{subfigure}
    \caption{ \add{Results of multi-GPU acceleration using 1 to 8 GPUs. Data is presented in two forms: execution time (left) and its inverse (right). The shaded regions in the figure represent the 95\% confidence interval of the data, but due to the dense data distribution, the shading is not noticeable.}}
    \label{fig:multi_cards}
\end{figure}

We also tested the performance of TensorNEAT in multi-GPU computing using the HalfCheetah environment. The population size was fixed at $5,000$, and each test involved running 50 generations of the NEAT algorithm. We conducted the experiments on Nvidia RTX 4090 hardware, varying the number of GPUs used (1, 2, 4, and 8). For each configuration, we averaged the total algorithm execution time over 10 independent runs to ensure reliable results. The results are presented in Fig.~\ref{fig:multi_cards}. The data is presented in two aspects: execution time and its inverse (execution performance). \add{It can be observed that, as the number of GPUs increases, execution time decreases significantly.} The improvement in execution performance is nearly linear, but the performance gains diminish when the number of GPUs increases from 4 to 8. This phenomenon is mainly attributed to two factors: 1. TensorNEAT employs multi-GPU acceleration only for the network evaluation phase, excluding the execution of the NEAT algorithm itself. As a result, only part of the overall execution time benefits from multi-GPU acceleration, leading to sub-linear performance improvements. 2. As the number of GPUs increases, TensorNEAT incurs additional overhead in managing communication between GPUs, which reduces the performance gain as more GPUs are utilized. \add{The multi-GPU experiments validate the effectiveness of TensorNEAT's acceleration while also demonstrating that its benefits tend to taper off as the number of GPUs increases.}

\subsection{Performance across Scalable Populations}

In this part, we investigate the impact of varying population sizes on the performance of TensorNEAT across three Brax environments: Swimmer, Hopper, and HalfCheetah. The primary objective is to assess how increasing the population size influences both the quality of the solutions and the convergence rate, thereby highlighting the benefits of the tensorization-based parallelism.

\begin{figure}[t]
    \centering    
    \begin{subfigure}{0.32\textwidth}
        \centering
        \includegraphics[width=\textwidth]{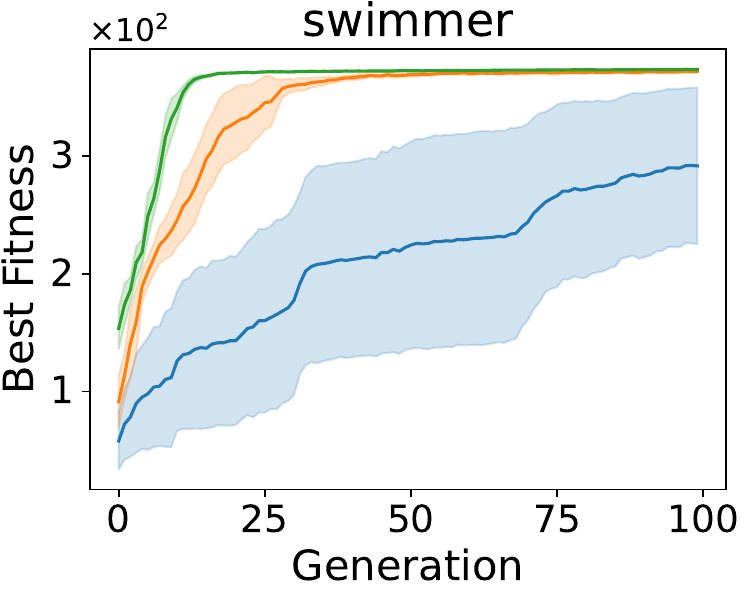}
    \end{subfigure}
    \begin{subfigure}{0.32\textwidth}
        \centering
        \includegraphics[width=\textwidth]{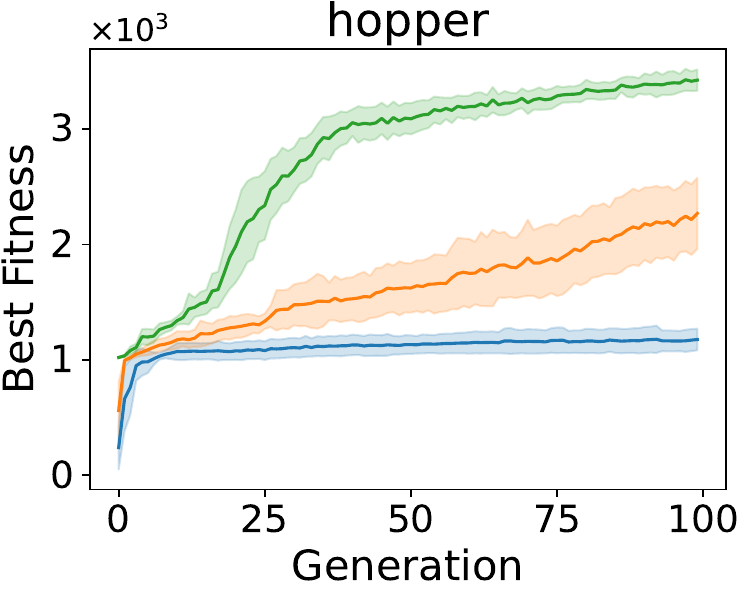}
    \end{subfigure}
    \begin{subfigure}{0.32\textwidth}
        \centering
        \includegraphics[width=\textwidth]{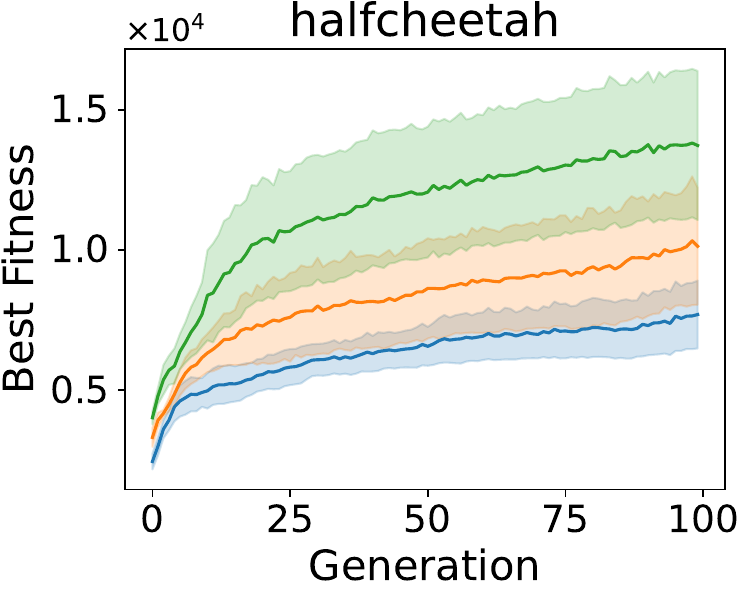}
    \end{subfigure}
    \begin{subfigure}{0.32\textwidth}
        \centering
        \includegraphics[width=\textwidth]{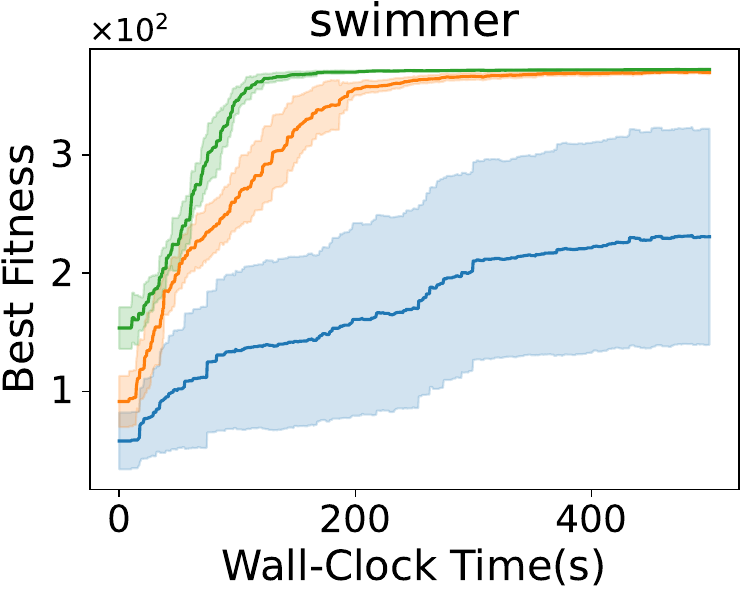}
    \end{subfigure}
    \begin{subfigure}{0.32\textwidth}
        \centering
        \includegraphics[width=\textwidth]{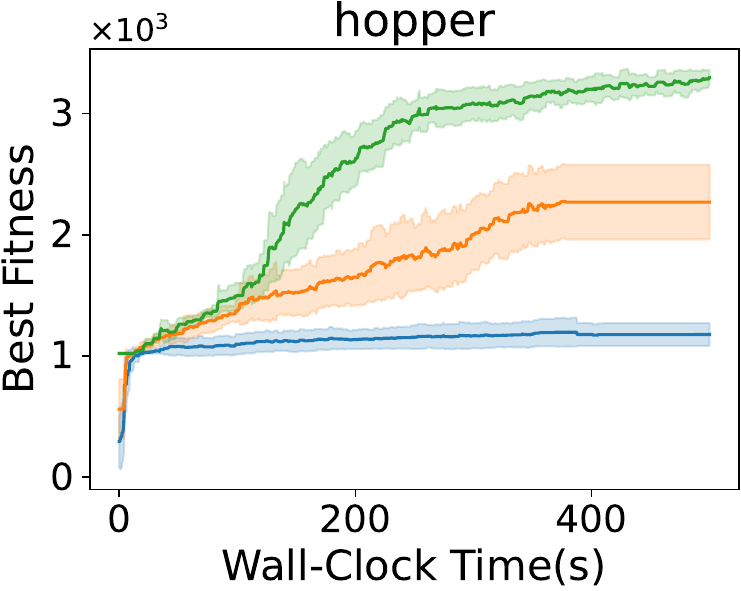}
    \end{subfigure}
    \begin{subfigure}{0.32\textwidth}
        \centering
        \includegraphics[width=\textwidth]{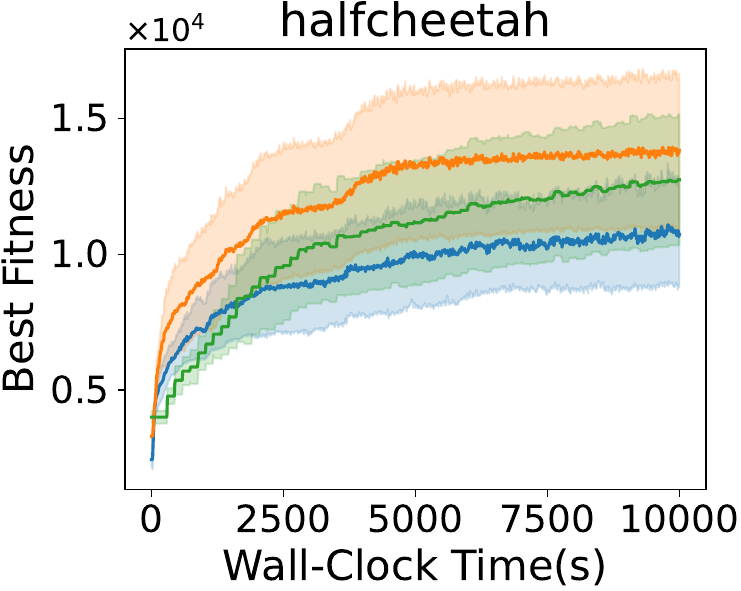}
    \end{subfigure}
    \begin{subfigure}{0.65\textwidth}
    \centering
    \includegraphics[width=\columnwidth]{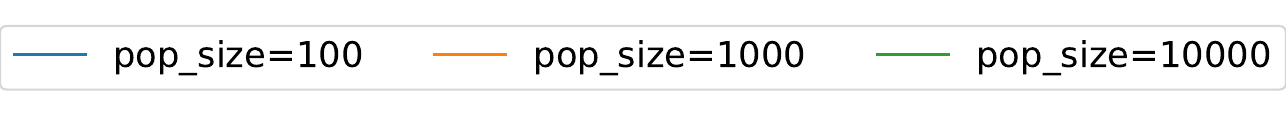}
    \end{subfigure}
    \caption{\add{Impact of population size on the best fitness achieved across Brax environments}}

    \label{fig:different-popsize}
\end{figure}

Fig.~\ref{fig:different-popsize} presents the experimental results, which are divided into two distinct parts. The top panel illustrates the best fitness achieved in the population after running a fixed number of generations for each population size. Across all three environments, larger population sizes consistently yield higher best fitness values. This result aligns with the intuition that a larger population enhances the search strength within each generation, leading to a more thorough exploration of the solution space.

The bottom panel of Fig.~\ref{fig:different-popsize} compares the best fitness attained after running for an equivalent wall-clock time. In most cases, larger population sizes reach higher fitness scores more rapidly. This observation demonstrates the effectiveness of TensorNEAT's parallel execution strategy. Unlike traditional serial implementations, where increasing the population size typically extends the computation time per generation, the tensorized approach significantly reduces this overhead. As a result, larger populations can sustain a high optimization rate while remaining within the hardware limits.

However, the HalfCheetah environment exhibits a notable exception. In this scenario, a population size of 10,000 results in a lower optimization rate compared to a population size of 1,000. This outcome is attributed to the increased complexity of the HalfCheetah task, which demands computational resources beyond the hardware's parallel processing capacity. Consequently, when the population size exceeds the hardware-supported limits, the benefits of parallelism diminish, leading to longer execution times per generation and a subsequent reduction in the optimization rate.

Overall, the parallelism introduced by tensorization not only accelerates the algorithm's execution but also enables the use of larger population sizes to enhance optimization efficiency. However, the performance gains from larger populations are constrained by the complexity of the problem and the hardware's parallel processing capacity.

\subsection{Comparison with evosax}

In this set of experiments, we compare TensorNEAT with evosax~\cite{lange2022evosax}, a JAX-based GPU-accelerated evolutionary algorithm library. We selected several GA variants available in evosax, including SimpleGA~\cite{simple-GA}, SAMR-GA~\cite{SAMR-GA}, LGA~\cite{LGA}, and GESMR-GA~\cite{GESMR-GA}, to benchmark against the NEAT algorithm implemented in TensorNEAT. For evosax, the default hyperparameter settings were used, and the policy network was instantiated as a four-layer MLP with 16 neurons in each hidden layer (\texttt{[inputs->hidden->hidden->outputs]}). In all experiments, the population size was fixed at 5000 and the maximum number of generations was set to 100.

\begin{figure}[t]
    \centering    
    \begin{subfigure}{0.32\textwidth}
        \centering
        \includegraphics[width=\textwidth]{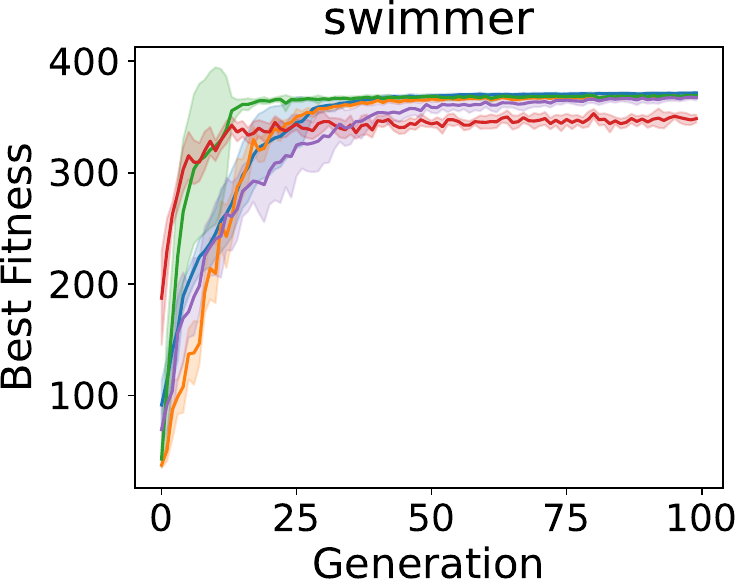}
    \end{subfigure}
    \begin{subfigure}{0.32\textwidth}
        \centering
        \includegraphics[width=\textwidth]{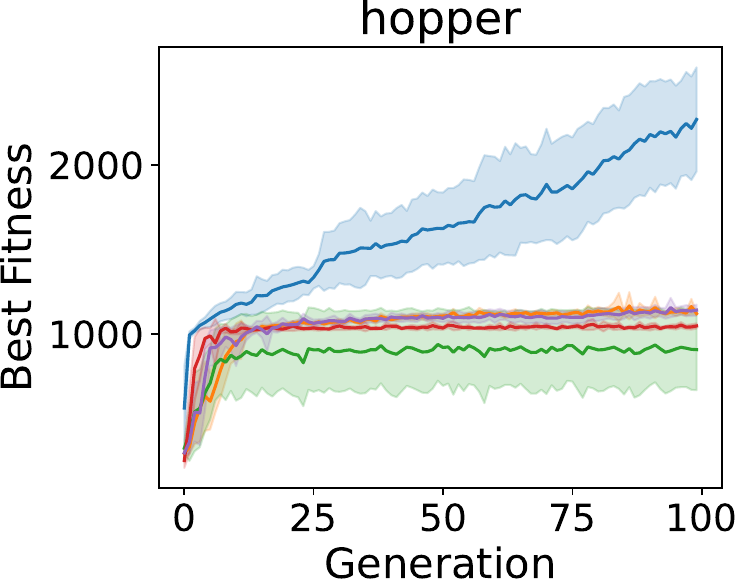}
    \end{subfigure}
    \begin{subfigure}{0.32\textwidth}
        \centering
        \includegraphics[width=\textwidth]{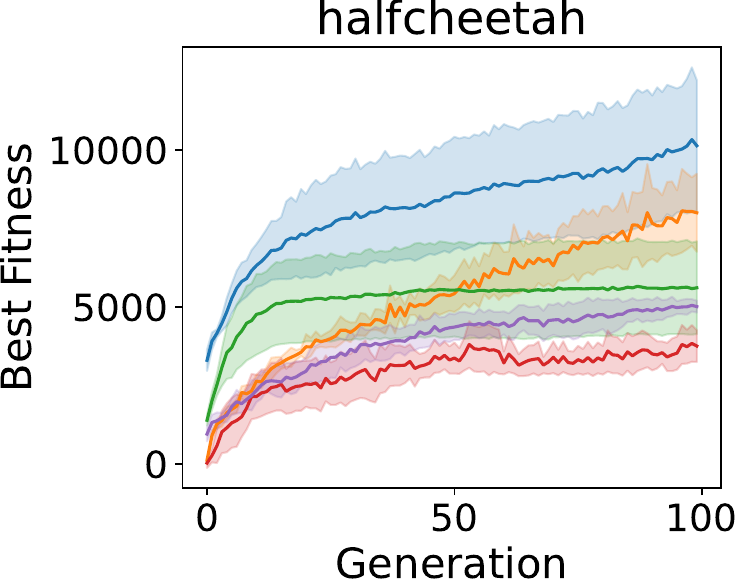}
    \end{subfigure}
    \begin{subfigure}{0.65\textwidth}
    \centering
    \includegraphics[width=\columnwidth]{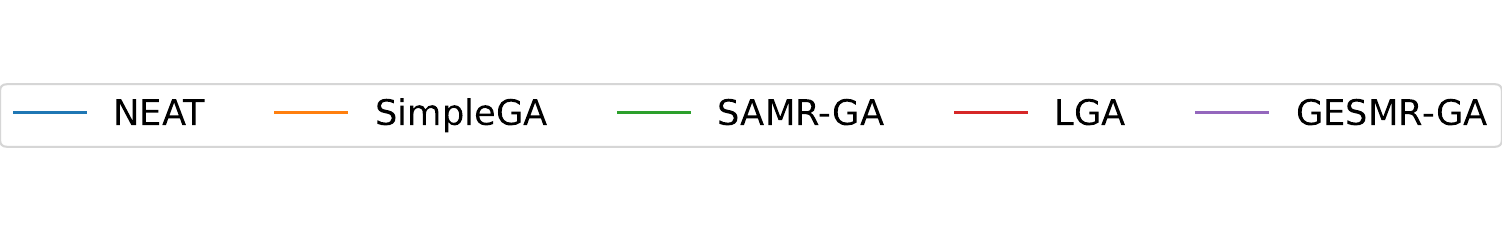}
    \end{subfigure}
    \caption{\add{Best fitness trends across Brax environments}}
    \label{fig:different-ga}
\end{figure}

Fig.~\ref{fig:different-ga} and Table~\ref{table:evosax} summarize the experimental outcomes. On the relatively simple Swimmer task, the NEAT algorithm achieved performance comparable to that of the GA variants in evosax, indicating that both methods are effective in less complex environments. However, in the more challenging Hopper and Halfcheetah tasks, the NEAT algorithm consistently outperformed all GA variants. These results highlight the advantages of NEAT in robotic control tasks, particularly its ability to evolve network architectures that are better suited to the demands of complex environments.

The superior performance of the NEAT algorithm can be attributed to its incremental growth mechanism, which begins with minimal network structures and progressively expands the topology. This gradual expansion of the search space allows the algorithm to adaptively scale its complexity in accordance with the task requirements. Additionally, the speciation mechanism inherent in NEAT preserves promising network structures, further facilitating effective exploration of the solution space.

It is noteworthy that TensorNEAT incurred an approximate 10\% increase in execution time relative to the GA variants in evosax across all tasks. We attribute this overhead to differences in network architectures. Specifically, the NEAT algorithm employs networks with irregular and randomly generated topologies, necessitating the individual computation of each node's value. This process involves numerous memory read and write operations on the GPU, which are less efficient compared to the streamlined matrix multiplication operations used in the MLPs of the GA variants. Despite this additional computational cost, the enhanced performance of NEAT in more complex tasks justifies the marginal increase in runtime.

In summary, the experiments compared to evosax demonstrate that although the NEAT algorithm requires slightly more computation time, its adaptive network growth and speciation mechanisms deliver superior performance in challenging robotic control tasks. These findings validate the effectiveness of our GPU-accelerated implementation and underscore the practical merit of TensorNEAT.

\begin{table*}[htbp]
    \centering
    \caption{\add{Performance and runtime comparison over $100$ generations across tasks}}
    \begin{tabular}{c|c|c|c|c}
        \hline
        Task & Library & Algorithm & Performance & Cost Time(s) \\
        \hline
        \multirow{5}{*}{Swimmer} 
            & TensorNEAT & NEAT       & \bm{$372.06 \pm 0.87$}   & $731.02 \pm 60.25$ \\
        \cline{2-5}
            & \multirow{4}{*}{evosax} 
            & SimpleGA   & $370.97 \pm 0.55$   & $685.37 \pm 45.15$ \\
            &            & SAMR-GA    & $371.24 \pm 0.82$   & $721.35 \pm 74.82$ \\
            &            & LGA        & $361.68 \pm 2.14$   & $726.42 \pm 82.89$ \\
            &            & GESMR-GA   & $369.41 \pm 2.21$   & \bm{$641.73 \pm 74.06$} \\
        \hline
        \multirow{5}{*}{Hopper} 
            & TensorNEAT & NEAT       & \bm{$2353.97 \pm 308.07$} & $364.72 \pm 7.47$ \\
        \cline{2-5}
            & \multirow{4}{*}{evosax} 
            & SimpleGA   & $1293.68 \pm 89.78$  & $304.16 \pm 2.96$ \\
            &            & SAMR-GA    & $983.02 \pm 178.11$  & $320.26 \pm 7.81$ \\
            &            & LGA        & $1140.61 \pm 16.80$  & \bm{$303.34 \pm 1.80$} \\
            &            & GESMR-GA   & $1194.43 \pm 42.63$  & $315.27 \pm 11.49$ \\
        \hline
        \multirow{5}{*}{Halfcheetah} 
            & TensorNEAT & NEAT       & \bm{$10481.12 \pm 2315.80$} & $1431.43 \pm 13.94$ \\
        \cline{2-5}
            & \multirow{4}{*}{evosax} 
            & SimpleGA   & $8748.52 \pm 1441.24$  & $1407.41 \pm 2.13$ \\
            &            & SAMR-GA    & $5776.35 \pm 1517.56$  & \bm{$1406.16 \pm 2.05$} \\
            &            & LGA        & $4705.10 \pm 554.95$   & $1406.41 \pm 2.16$ \\
            &            & GESMR-GA   & $5275.45 \pm 310.59$   & $1408.14 \pm 2.32$ \\
        \hline
    \end{tabular}
    \label{table:evosax}
\end{table*}

\section{Conclusion}
In this paper, we tackled the scalability limitations of the NeuroEvolution of Augmenting Topologies (NEAT) algorithm by introducing a tensorization approach, which converts network topologies into uniformly shaped tensors for efficient parallel computation. Building on this foundation, we developed TensorNEAT, a library that utilizes JAX for automatic function vectorization and hardware acceleration, enabling seamless execution on modern GPUs and TPUs. TensorNEAT's flexible design allows for easy integration with widely-used reinforcement learning environments such as Gym, Brax, and gymnax, supporting extensive experimentation in robotics and control tasks. Our method not only simplifies the parallelization of the evolutionary process but also significantly reduces the computational cost of large-scale simulations. Consequently, TensorNEAT achieves up to 500x speedups compared to conventional NEAT implementations, establishing it as a powerful tool for high-performance neuroevolution, especially in scenarios requiring real-time feedback or complex multi-agent interactions. This substantial improvement in computational efficiency paves the way for researchers to explore more advanced evolutionary algorithms and tackle more challenging real-world applications.

Our findings add to the current understanding and may pave the way for future research in this field. Looking ahead, our roadmap for TensorNEAT includes expanding its reach to distributed computing environments, transcending the limitations of single-machine setups. By enabling TensorNEAT to operate in distributed architectures, we aim to significantly improve scalability, allowing the system to handle larger datasets and more complex neural network topologies with greater efficiency. Furthermore, we plan to augment TensorNEAT's suite of functionalities by integrating advanced NEAT variants, such as DeepNEAT and CoDeepNEAT \cite{miikkulainen2019evolving}, to further enhance its potential in solving complex neuroevolution challenges. These variants introduce hierarchical and modular structures that can better exploit the representational power of deep learning models, potentially accelerating both the convergence speed and the quality of evolved solutions. Our long-term goal is to ensure TensorNEAT remains adaptable and robust, providing researchers and practitioners with a versatile tool for tackling a diverse range of evolutionary computation problems across various domains.

\bibliographystyle{ACM-Reference-Format}
\bibliography{neatax-reference}

%
\appendix
\newpage
\onecolumn{}
\section{Hyperparameters}\label{Appendix_a}

The hyperparameters in TensorNEAT consists of those that influence the algorithm's dynamics and those who affect network behaviors:

\begin{itemize}
    \item Algorithmic Controls:
    \begin{itemize}
        \item \texttt{seed}: Random seed (integer).
        \item \texttt{fitness\_target}: Target fitness value for termintaion (float).
        \item \texttt{generation\_limit}: Maximum number of generations for termintaion (float).
        \item \texttt{pop\_size}: Population size (integer).
        \item \texttt{network\_type}: Network type, either \texttt{feedforward} (no cycles) or \texttt{recurrent} (with cycles).
        \item \texttt{inputs}: Number of network inputs (integer).
        \item \texttt{outputs}: Number of network outputs (integer).
        \item \texttt{max\_nodes}: Max nodes allowed in a network (integer).
        \item \texttt{max\_conns}: Max connections allowed in a network (integer).
        \item \texttt{max\_species}: Max species in the population (integer).
        \item \texttt{compatibility\_disjoint}: Weight for disjoint genes in distance calculation between genomes (float).
        \item \texttt{compatibility\_homologous}: Weight for homologous genes in distance calculation between genomes (float).
        \item \texttt{node\_add}: Probability of a node addition during mutation (float).
        \item \texttt{node\_delete}: Probability of a node deletion during mutation (float).
        \item \texttt{conn\_add}: Probability of a connection addition during mutation (float).
        \item \texttt{conn\_delete}: Probability of a connection deletion during mutation (float).
        \item \texttt{compatibility\_threshold}: Distance threshold for genome speciation (float). 
        \item \texttt{species\_elitism}: Minimum species count to prevent all species are stagnated (integer).
        \item \texttt{max\_stagnation}: Stagnation threshold for species (integer). If a species does not show improvement for \texttt{max\_stagnation} consecutive generations, then this species will be stagnated.
        \item \texttt{genome\_elitism}: Number of elite genomes preserved for next generation (integer).
        \item \texttt{survival\_threshold}: Percentage of species survival for crossover (float).
        \item \texttt{spawn\_number\_change\_rate}: Rate of change in species size over two consecutive generations (float).
    \end{itemize}

    \item Network Behavior Controls:
    \begin{itemize}
        \item \texttt{bias\_init\_mean}: Mean value for bias initialization (float).
        \item \texttt{bias\_init\_std}: Standard deviation for bias initialization (float).
        \item \texttt{bias\_mutate\_power}: Mutation strength for bias values (float).
        \item \texttt{bias\_mutate\_rate}: Probability of bias value mutation (float).
        \item \texttt{bias\_replace\_rate}: Probability to replace existing bias with a new value during mutation (float).

        \item \texttt{response\_init\_mean}: Mean value for response initialization (float).
        \item \texttt{response\_init\_std}: Standard deviation for response initialization (float).
        \item \texttt{response\_mutate\_power}: Mutation strength for response values (float).
        \item \texttt{response\_mutate\_rate}: Probability of response value mutation (float).
        \item \texttt{response\_replace\_rate}: Probability to replace existing response with a new value during mutation (float).

        \item \texttt{weight\_init\_mean}: Mean value for weight initialization (float).
        \item \texttt{weight\_init\_std}: Standard deviation for weight initialization (float).
        \item \texttt{weight\_mutate\_power}: Mutation strength for weight values (float).
        \item \texttt{weight\_mutate\_rate}: Probability of weight value mutation (float).
        \item \texttt{weight\_replace\_rate}: Probability to replace existing weight with a new value during mutation (float).

        \item \texttt{activation\_default}: Default value for activation function.
        \item \texttt{activation\_options}: Available activation functions.
        \item \texttt{activation\_replace\_rate}: Probability to change the activation function during mutation.

        \item \texttt{aggregation\_default}: Default value for aggregation function.
        \item \texttt{aggregation\_options}: Available aggregation functions.
        \item \texttt{aggregation\_replace\_rate}: Probability to change the aggregation function during mutation.
    \end{itemize}
\end{itemize}

\newpage
\section{Interfaces}\label{Appendix_b}

\subsection{Network Interface}

TensorNEAT offers users the flexibility to define a custom network that will be optimized by the NEAT algorithms by providing a network interface (as presented in Listing~\ref{lst:gene_interface} and Listing~\ref{lst:genome_interface}). Users are required to define the specific behaviors associated with their network, including initialization, mutation, distance computation, and inference.

By implementing these interfaces, users can fit their specific requirements and leverage TensorNEAT's power for a wide range of neural network architectures.

\begin{figure}[h]
\begin{lstlisting}[
language=Python,
caption=Gene interface. ,
label=lst:gene_interface
]
class BaseGene:
    fixed_attrs = []
    custom_attrs = []

    def __init__(self):
        pass

    def new_custom_attrs(self):
        raise NotImplementedError

    def mutate(self, randkey, gene):
        raise NotImplementedError

    def distance(self, gene1, gene2):
        raise NotImplementedError

    def forward(self, attrs, inputs):
        raise NotImplementedError

    @property
    def length(self):
        return len(self.fixed_attrs) + len(self.custom_attrs)
\end{lstlisting}
\end{figure}

\begin{figure}[h]
\begin{lstlisting}[
language=Python,
caption=Genome interface. ,
label=lst:genome_interface
]
class BaseGenome:
    network_type = None

    def __init__(
            self,
            num_inputs: int,
            num_outputs: int,
            max_nodes: int,
            max_conns: int,
            node_gene: BaseNodeGene = DefaultNodeGene(),
            conn_gene: BaseConnGene = DefaultConnGene(),
    ):
        self.num_inputs = num_inputs
        self.num_outputs = num_outputs
        self.input_idx = jnp.arange(num_inputs)
        self.output_idx = jnp.arange(num_inputs, num_inputs + num_outputs)
        self.max_nodes = max_nodes
        self.max_conns = max_conns
        self.node_gene = node_gene
        self.conn_gene = conn_gene

    def transform(self, nodes, conns):
        raise NotImplementedError

    def forward(self, inputs, transformed):
        raise NotImplementedError
\end{lstlisting}
\end{figure}

\subsection{Problem Template}

In TensorNEAT, users also have the flexibility to define custom problems they wish to optimize using the NEAT algorithms. This can be done by implementing the \texttt{Problem} interface as illustrated in Listing~\ref{lst:problem_interface}. Within this interface, users are required to provide the evaluation process for the problem, specify the input and output dimensions, and optionally implement a function to visualize the solution.

\begin{figure}[H]
\begin{lstlisting}[
language=Python,
caption=Problem interface in TensorNEAT. ,
label=lst:problem_interface
]
from typing import Callable

from utils import State


class BaseProblem:
    jitable = None

    def setup(self, randkey, state: State = State()):
        """initialize the state of the problem"""
        raise NotImplementedError

    def evaluate(self, randkey, state: State, act_func: Callable, params):
        """evaluate one individual"""
        raise NotImplementedError

    @property
    def input_shape(self):
        raise NotImplementedError

    @property
    def output_shape(self):
        raise NotImplementedError

    def show(self, randkey, state: State, act_func: Callable, params, *args, **kwargs):
        raise NotImplementedError

\end{lstlisting}
\end{figure}

\section{Experiment Detail}\label{Appendix_c}
In this section, we detail the hyperparameters in experiments.

\subsection{Hyperparameters}

Hyperparameters in TensorNEAT:
\begin{itemize}
    \item Algorithmic Controls:
    \begin{itemize}
        \item \texttt{seed}: \texttt{[0, 1, 2, 3, 4, 5, 6, 7, 8, 9]};
        \item \texttt{fitness\_target}: \texttt{Inf (to terminate at the fixed generation)};
        \item \texttt{generation\_limit}: \texttt{100};
        \item \texttt{pop\_size}: \texttt{[50, 100, 200, 500, 1000, 2000, 5000, 10000]};
        \item \texttt{network\_type}: \texttt{feedforward};
        \item \texttt{inputs}: the same as the observation dimension of the environment.
        \item \texttt{outputs}: the same as the action dimension of the environment.
        \item \texttt{max\_nodes}: \texttt{50};
        \item \texttt{max\_conns}: \texttt{100};
        \item \texttt{max\_species}: \texttt{10};
        \item \texttt{compatibility\_disjoint}: \texttt{1.0};
        \item \texttt{compatibility\_homologous}: \texttt{0.5};
        \item \texttt{node\_add}: \texttt{0.2};
        \item \texttt{node\_delete}: \texttt{0};
        \item \texttt{conn\_add}: \texttt{0.4};
        \item \texttt{conn\_delete}: \texttt{0};
        \item \texttt{compatibility\_threshold}: \texttt{3.5};
        \item \texttt{species\_elitism}: \texttt{2};
        \item \texttt{max\_stagnation}: \texttt{15};
        \item \texttt{genome\_elitism}: \texttt{2};
        \item \texttt{survival\_threshold}: \texttt{0.2};
        \item \texttt{spawn\_number\_change\_rate}: \texttt{0.5};
    \end{itemize}

    \item Network Behavior Controls:
    \begin{itemize}
        \item \texttt{bias\_init\_mean}: \texttt{0};
        \item \texttt{bias\_init\_std}: \texttt{1.0};
        \item \texttt{bias\_mutate\_power}: \texttt{0.5};
        \item \texttt{bias\_mutate\_rate}: \texttt{0.7};
        \item \texttt{bias\_replace\_rate}: \texttt{0.1};

        \item \texttt{response\_init\_mean}: \texttt{1.0};
        \item \texttt{response\_init\_std}: \texttt{0};
        \item \texttt{response\_mutate\_power}: \texttt{0};
        \item \texttt{response\_mutate\_rate}: \texttt{0};
        \item \texttt{response\_replace\_rate}: \texttt{0};

        \item \texttt{weight\_init\_mean}: \texttt{0};
        \item \texttt{weight\_init\_std}: \texttt{1};
        \item \texttt{weight\_mutate\_power}: \texttt{0.5};
        \item \texttt{weight\_mutate\_rate}: \texttt{0.8};
        \item \texttt{weight\_replace\_rate}: \texttt{0.1};

        \item \texttt{activation\_default}: \texttt{tanh};
        \item \texttt{activation\_options}: \texttt{[tanh]};
        \item \texttt{activation\_replace\_rate}: \texttt{0};

        \item \texttt{aggregation\_default}: \texttt{sum};
        \item \texttt{aggregation\_options}: \texttt{[sum]};
        \item \texttt{aggregation\_replace\_rate}: \texttt{0};
    \end{itemize}
\end{itemize}

Hyperparameters in NEAT-Python:
\begin{itemize}
    \item \texttt{fitness\_criterion}: \texttt{max}
    \item \texttt{fitness\_threshold}: \texttt{999}
    \item \texttt{pop\_size}: \texttt{10000}
    \item \texttt{reset\_on\_extinction}: \texttt{False}

    \item \textbf{[DefaultGenome]}
    \begin{itemize}
        \item \texttt{activation\_default}: \texttt{tanh}
        \item \texttt{activation\_mutate\_rate}: \texttt{0}
        \item \texttt{activation\_options}: \texttt{tanh}
        \item \texttt{aggregation\_default}: \texttt{sum}
        \item \texttt{aggregation\_mutate\_rate}: \texttt{0.0}
        \item \texttt{aggregation\_options}: \texttt{sum}
        \item \texttt{bias\_init\_mean}: \texttt{0.0}
        \item \texttt{bias\_init\_stdev}: \texttt{1.0}
        \item \texttt{bias\_max\_value}: \texttt{30.0}
        \item \texttt{bias\_min\_value}: \texttt{-30.0}
        \item \texttt{bias\_mutate\_power}: \texttt{0.5}
        \item \texttt{bias\_mutate\_rate}: \texttt{0.7}
        \item \texttt{bias\_replace\_rate}: \texttt{0.1}
        \item \texttt{compatibility\_disjoint\_coefficient}: \texttt{1.0}
        \item \texttt{compatibility\_weight\_coefficient}: \texttt{0.5}
        \item \texttt{conn\_add\_prob}: \texttt{0.5}
        \item \texttt{conn\_delete\_prob}: \texttt{0}
        \item \texttt{enabled\_default}: \texttt{True}
        \item \texttt{enabled\_mutate\_rate}: \texttt{0.01}
        \item \texttt{feed\_forward}: \texttt{True}
        \item \texttt{initial\_connection}: \texttt{full}
        \item \texttt{node\_add\_prob}: \texttt{0.2}
        \item \texttt{node\_delete\_prob}: \texttt{0}
        \item \texttt{num\_hidden}: \texttt{0}
        \item \texttt{num\_inputs}: the same as the observation dimension of the environment.
        \item \texttt{num\_outputs}: the same as the action dimension of the environment.
        \item \texttt{response\_init\_mean}: \texttt{1.0}
        \item \texttt{response\_init\_stdev}: \texttt{0.0}
        \item \texttt{response\_max\_value}: \texttt{30.0}
        \item \texttt{response\_min\_value}: \texttt{-30.0}
        \item \texttt{response\_mutate\_power}: \texttt{0.0}
        \item \texttt{response\_mutate\_rate}: \texttt{0.0}
        \item \texttt{response\_replace\_rate}: \texttt{0.0}
        \item \texttt{weight\_init\_mean}: \texttt{0.0}
        \item \texttt{weight\_init\_stdev}: \texttt{1.0}
        \item \texttt{weight\_max\_value}: \texttt{30}
        \item \texttt{weight\_min\_value}: \texttt{-30}
        \item \texttt{weight\_mutate\_power}: \texttt{0.5}
        \item \texttt{weight\_mutate\_rate}: \texttt{0.8}
        \item \texttt{weight\_replace\_rate}: \texttt{0.1}
    \end{itemize}

    \item \textbf{[DefaultSpeciesSet]}
    \begin{itemize}
        \item \texttt{compatibility\_threshold}: \texttt{3.0}
    \end{itemize}

    \item \textbf{[DefaultStagnation]}
    \begin{itemize}
        \item \texttt{species\_fitness\_func}: \texttt{max}
        \item \texttt{max\_stagnation}: \texttt{20}
        \item \texttt{species\_elitism}: \texttt{2}
    \end{itemize}

    \item \textbf{[DefaultReproduction]}
    \begin{itemize}
        \item \texttt{elitism}: \texttt{-9999}
        \item \texttt{survival\_threshold}: \texttt{0.2}
    \end{itemize}
\end{itemize}

Hyperparameters in evosax:

\begin{itemize}
    \item \textbf{[SimpleGA]}
    \begin{itemize}
        \item \textbf{cross\_over\_rate}: 0.0
        \item \textbf{sigma\_init}: 0.07
        \item \textbf{sigma\_decay}: 1.0
        \item \textbf{sigma\_limit}: 0.01
        \item \textbf{init\_min}: 0.0
        \item \textbf{init\_max}: 0.0
        \item \textbf{clip\_min}: $-\text{jnp.finfo(jnp.float32).max}$
        \item \textbf{clip\_max}: $\text{jnp.finfo(jnp.float32).max}$
        \item \textbf{elite\_ratio}: 0.5
    \end{itemize}

    \item \textbf{[SAMR-GA]}
    \begin{itemize}
        \item \textbf{sigma\_init}: 0.07
        \item \textbf{sigma\_meta}: 2.0
        \item \textbf{sigma\_best\_limit}: 0.0001
        \item \textbf{init\_min}: 0.0
        \item \textbf{init\_max}: 0.0
        \item \textbf{clip\_min}: $-\text{jnp.finfo(jnp.float32).max}$
        \item \textbf{clip\_max}: $\text{jnp.finfo(jnp.float32).max}$
        \item \textbf{elite\_ratio}: 0.0
    \end{itemize}

    \item \textbf{[LGA]}
    \begin{itemize}
        \item \textbf{cross\_over\_rate}: 0.0
        \item \textbf{sigma\_init}: 1.0
        \item \textbf{init\_min}: -5.0
        \item \textbf{init\_max}: 5.0
        \item \textbf{clip\_min}: $-\text{jnp.finfo(jnp.float32).max}$
        \item \textbf{clip\_max}: $\text{jnp.finfo(jnp.float32).max}$
        \item \textbf{elite\_ratio}: 1.0
    \end{itemize}

    \item \textbf{[GESMR-GA]}
    \begin{itemize}
        \item \textbf{sigma\_init}: 0.07
        \item \textbf{sigma\_meta}: 2.0
        \item \textbf{init\_min}: 0.0
        \item \textbf{init\_max}: 0.0
        \item \textbf{clip\_min}: $-\text{jnp.finfo(jnp.float32).max}$
        \item \textbf{clip\_max}: $\text{jnp.finfo(jnp.float32).max}$
        \item \textbf{elite\_ratio}: 0.5
        \item \textbf{sigma\_ratio}: 0.5
    \end{itemize}

\end{itemize}

\end{document}